%% file: binaryConf-arXiv.tex
\documentclass{article}

\usepackage{arxiv}

\usepackage[utf8]{inputenc} 
\usepackage[T1]{fontenc}    
\usepackage{hyperref}       
\usepackage{url}            
\usepackage{booktabs}       
\usepackage{amsfonts}       
\usepackage{nicefrac}       
\usepackage{microtype}      
\usepackage{lipsum}		    
\usepackage{graphicx}
\usepackage{natbib}[numbers]
\usepackage{doi}

\usepackage{amsmath}
\usepackage{amssymb}
\usepackage[parindent=\parindent]{subcaption}
\usepackage{fancyvrb}
\usepackage{makecell}
\usepackage{multirow}
\usepackage[most]{tcolorbox}
\usepackage{placeins}
\usepackage[leftcaption]{sidecap}
\sidecaptionvpos{figure}{t}
\usepackage{mathtools}
\usepackage{blkarray}
\usepackage{pbox}
\usepackage{ragged2e}
\usepackage[most]{tcolorbox}
\usepackage{mathtools}
\newcommand\SmallMatrix[1]{{\tiny\arraycolsep=0.3\arraycolsep\ensuremath{\begin{bmatrix}#1\end{bmatrix}}}}

\allowdisplaybreaks

\newlength{\savedparindent}
\AtBeginDocument{\setlength{\savedparindent}{\parindent}}
\hypersetup{colorlinks=true,allcolors=blue}
\input{newcommands}

\title{Never mind the metrics---what about the uncertainty?\\
Visualising confusion matrix metric distributions}

\author{ \href{https://orcid.org/0000-0002-3938-7586}{\includegraphics[scale=0.06]{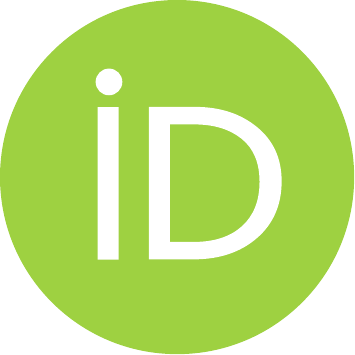}\hspace{1mm}David Lovell}\\
	Queensland University of Technology\\
	Centre for Data Science and\\
	School of Computer Science\\
	Brisbane, Australia \\
	\texttt{David.Lovell@qut.edu.au} \\
	\And
	\href{https://orcid.org/0000-0001-6312-8325}{\includegraphics[scale=0.06]{orcid.pdf}\hspace{1mm}Dimity Miller} \\
	Queensland University of Technology\\
	School of Computer Science\\
	Brisbane, Australia \\
	\texttt{d24.miller@qut.edu.au} \\
	\And
	{Jaiden Capra} \\
	Queensland University of Technology\\
	Brisbane, Australia \\
	\texttt{Jaiden.Capra@connect.qut.edu.au} \\
	\And
	\href{https://orcid.org/0000-0003-0109-6844}{\includegraphics[scale=0.06]{orcid.pdf}\hspace{1mm}Andrew P. Bradley} \\
	Queensland University of Technology\\
	Brisbane, Australia \\
	\texttt{a6.Bradley@qut.edu.au}
}

\renewcommand{\headeright}{Technical Report}
\renewcommand{\undertitle}{Technical Report}
\renewcommand{\shorttitle}{Visualising confusion matrix metric distributions}

\hypersetup{
pdftitle={Never mind the metrics---what about the uncertainty?
Visualising confusion matrix metric distributions},
pdfsubject={classification performance measurement},
pdfauthor={David Lovell, Dimity Miller, Jaiden Capra, Andrew Bradley},
pdfkeywords={Confusion matrix, performance metric, class imbalance, beta-binomial distribution, ROC},
}

\begin{document}

\maketitle

\begin{abstract}
    There are strong incentives to build models that demonstrate outstanding predictive performance on various datasets and benchmarks. We believe these incentives risk a narrow focus on models and on the performance metrics used to evaluate and compare them---resulting in a growing body of literature to evaluate and compare \textit{metrics}. This paper strives for a more balanced perspective on classifier performance metrics by highlighting their distributions under different models of uncertainty and showing how this uncertainty can easily eclipse differences in the empirical performance of classifiers. We begin by emphasising the fundamentally discrete nature of empirical confusion matrices and show how binary matrices can be meaningfully represented in a three dimensional compositional lattice, whose cross-sections form the basis of the space of receiver operating characteristic (ROC) curves. We develop equations, animations and interactive visualisations of the contours of performance metrics within (and beyond) this ROC space, showing how some are affected by class imbalance. We provide interactive visualisations that show the discrete posterior predictive probability mass functions of true and false positive rates in ROC space, and how these relate to uncertainty in performance metrics such as Balanced Accuracy (BA) and the Matthews Correlation Coefficient (MCC). Our hope is that these insights and visualisations will raise greater awareness of the substantial uncertainty in performance metric estimates that can arise when classifiers are evaluated on empirical datasets and benchmarks, and that classification model performance claims should be tempered by this understanding.
\end{abstract}


\section{Introduction}

Today's algorithmic modeling culture \citep{breiman_statistical_2001} venerates and rewards individuals whose predictive models outperform all others. Performance optimisation is central to statistical, machine learning and AI models \citep{thomas_problem_2020}, and numerical metrics are regarded by many as objective, valid indicators of performance; a model's empirical performance on a test dataset or benchmark task is seen as an indication of its ability to perform ``in the wild'', i.e., on unseen data obtained from real-world application. This paper strives for a more balanced perspective on classifier performance metrics by highlighting, visualising and providing insight into their distributions under different models of uncertainty. We aim to draw people's attention towards the role of more data---and data that is more \textit{representative} of a classifier's intended application---to characterise and understand the strengths and limitations of different classifiers, rather than judging them through performance shoot-outs alone.

Confusion matrices are a popular way to summarise the empirical performance of classifiers. Simplest are the confusion matrices that tally the four possible outcomes of \textit{binary decisions} (Table~\ref{tab:confusion-matrix}). Still, a desire to further summarise each matrix with one number---which, necessarily loses information because the matrices have three degrees of freedom---has led to a diverse menagerie of performance metrics (Appendix~\ref{app:metrics}). These metrics have different meanings, interpretations and ranges but, generally speaking, the bigger the better: the larger the value of a performance metric, the better the classifier has performed on the data it has been given.

However, different performance matrices produce different classifier performance rankings, and there are often strong reputational and financial incentives to be at the top of these ranks \citep{maier-hein_why_2018}. Those seeking to develop or promote ``the best classifier'' may wonder which is ``the best performance metric'', and there have been several studies arguing the relative merits of different metrics, even though what is ``best'' in practice depends on the specifics of a classifier's real-world application \citep{hand_classifier_2006,rudin_why_2019}. Others have sought to understand and characterise the behaviour and interpretation of these different metrics; the work we present here follows in their philosophical footsteps to reveal further insight.

Specifically, we emphasise the fundamentally discrete nature of binary empirical confusion matrices and show how their four counts can be well represented using a three-dimensional compositional lattice. We present visualisations and animations that demonstrate how different performance metrics relate to slices of that lattice corresponding to specific numbers of positive and negative examples. We derive and visualise the contours of these performance metrics in terms of true positive and true negative rates and reveal their geometry both inside and outside the space of ROC curves.
We show that confusion matrix performance metrics take on discrete values and provide a Bayesian approach to characterising the distributions (i.e., probability mass functions) of different metrics based on empirical observations (i.e., observed confusion matrices). This is important because it reminds us that there can be considerable uncertainty in our estimates of classifier performance, and that performance claims should be tempered by that fact.

In essence, this paper uses visualisation to help understand how uncertainty in (discrete) confusion matrices manifests in various (continuous) performance metrics so that practitioners can put classifier performance estimates and comparisons into perspective. Our hope is that this will encourage more attention towards reducing uncertainty in performance estimates before attempting to argue the merits or limitations of a particular classifier or metric.

The work we present here is motivated by our desire to make sense of confusion matrices and has been shaped, inspired and provoked by a range of previous investigations which we now review.

\begin{SCtable}
\caption{A binary confusion matrix shows the counts of a classifier's predictions in response to a set of examples whose actual classes are known. As well as referring to these elements as true or false positives and negatives, we will label them $a, b, c, d$ to simplify formulae. When their total ($\Tot=a+b+c+d$) is fixed, these four-element matrices have three degrees of freedom. Single number performance metrics necessarily lose information.}
\makebox[0.6\textwidth]{\input{table.binary}}
\label{tab:confusion-matrix}
\end{SCtable}

\section{What prior studies motivate our investigation?}

Published research on performance metrics for binary classification is extensive and, for the purposes of this paper, can usefully be thought of under four broad, often overlapping themes. Here, we relate these themes to the knowledge gaps we address in this paper.

\subsection{Studies that argue the merits or limitations of one metric over another}

The research we present here was stimulated by a series of papers by Chicco and co-authors whose titles suggest that Matthews Correlation Coefficient (MCC, Equation~\eqref{eq:MCC}) is better than a range of other metrics \citep{chicco_advantages_2020,chicco_benefits_2021,chicco_matthews_2021}. While the story told by these papers is more nuanced than their titles suggest, their citation rates suggest that the clear message in their titles has caught people's attention, especially in bioinformatics, following promotion of MCC as a good performance measure for machine learning in computational biology \citep{chicco_ten_2017}. A recent opinion piece may see MCC's popularity rise in robotics and machine vision \citep{chicco_invitation_2022}.


\cite{zhu_performance_2020-1} challenges the idea that MCC should \textit{``generally regarded as a balanced measure which can be used even if the classes are of very different sizes,''} a quotation which appears to be from Wikipedia editors rather than peer-reviewed proponents of MCC (see Appendix~\ref{sec:MCCquote}). Note that \citet[Abstract]{chicco_matthews_2021} clearly state that MCC is perhaps not the best measure in all situations, including \textit{``analyzing classifications where dataset prevalence is unrepresentative, comparing classifiers on different datasets, and assessing the random guessing level of a classifier.''} They continue, however, to suggest that it be a standard measure for scientists of all fields.

In earlier work, \cite{glas_diagnostic_2003} make a carefully considered case for the diagnostic odds ratio (Equation~\eqref{eq:DOR}), pointing out its useful statistical properties in summarising the two likelihood ratios, $\LRP$ and $\LRN$, that relate the pre- and post-test probabilities of outcome $\PPV$ and $\NPV$ (Equations~\ref{eq:LRP},\ref{eq:LRN},\ref{eq:PPV},\ref{eq:NPV}). One important point to note is that diagnostic odds ratio has a clear probabilistic interpretation as the ratio of the odds of a true positive relative to the odds of a false positive---some authors seem to discount this and concentrate on DOR as a means only to compare classifier performance \citep{racz_multi-level_2019-1,chicco_benefits_2021}. Other works that focus on specific metrics include \cite{powers_what_2019} who critiques $F$-measures, and \cite{delgado_why_2019} who study Cohen's Kappa in relation to MCC.

Instead of arguing for or against specific metrics, this paper strives for a more balanced perspective by enabling practitioners to explore and understand the behaviour of different performance metrics and the uncertainty in their values that arises when they are estimated empirically. \textit{``Several rates that summarize the four categories of the confusion matrix exist nowadays; none of them, however, has reached consensus in the computer science''} \citep{chicco_benefits_2021}: we believe the idea that of consensus about ``the best'' performance metric  oversimplifies the fact that \textit{``each of the indicators serves a different purpose''} \citep{glas_diagnostic_2003}. Practitioners have a responsibility to understand the strengths and limitations of different indicators for the purpose at hand, and to appreciate that metrics only \textit{summarise} empirical performance; they do not themselves provide more precise estimates of performance---that demands more data that is representative of the context into which a classifier would be deployed.

\subsection{Studies that characterise, relate and compare a range of performance metrics}

Rather than advocating for or against specific performance metrics, some authors have sought to characterise and understand their behaviour and relationships. \cite{powers_evaluation_2011} describes algebraic relationships between several metrics before using Monte Carlo simulation to explore their behaviour. \cite{ferri_experimental_2009} look at the empirical behaviour of different metrics applied to a range of classification algorithms and datasets from the UCI Machine Learning Repository, then use cluster analysis to group metrics that behaved similarly. It is difficult to draw strong conclusions about metrics from the values they take on simulated or experimental data, especially when comparisons are averaged or clustered across different datasets.

Other authors seek to characterise performance metrics more directly by exploring how they satisfy various properties. \cite{sokolova_systematic_2009} consider how the values of different metrics change or remain constant as the counts in a confusion matrix change; it is important to note that their analysis compares metrics using confusion matrices with \textit{different} totals. In contrast, \cite{brzezinski_visual-based_2018} consider ten desirable properties of metrics on confusion matrices with \textit{fixed} totals, but different prevalence (class balances).
More recently, \cite{gosgens_good_2021} has considered further desirable properties of performance metrics and shown that three of these (``monotonicity'', ``distance'' and ``constant baseline'') are incompatible.

\cite{luque_impact_2019} focus on characterising the behaviour of different performance metrics as class balance changes. They express each performance metric as a function of true positive rate (Equation~\eqref{eq:TPR}), true negative rate (Equation~\eqref{eq:TNR}) and prevalence (Equation~\eqref{eq:Prev}) and treat each of these values as though they were continuous. (As we will emphasise later, these values are discrete because they are derived from integer counts.) They then explore and compare the distributions of different performance metrics as prevalence varies from 0 to 1, observing the similarities of these metrics in terms of various summary statistics.

Like \cite{brzezinski_visual-based_2018}, we aim to make the properties of different performance metrics more accessible to practitioners through interactive visualisation methods. We add to the repertoire of performance metric visualisation methods by presenting algebraic expressions for performance metric contours which clearly characterise their behaviour within and beyond ROC space as class imbalance changes.

\subsection{Studies that explore and visualise the structure and geometry associated with performance metrics}

Broadly speaking, there have been two main approaches to visualising the values that performances metrics take over the space of possible binary confusion matrices.

The first approach uses the \textit{rates} that characterise different  confusion matrices (e.g., true positive rate, false positive rate and prevalence \citep{flach_geometry_2003, luque_impact_2019}). This yields visualisations within the pseudo-cube $[0,1]\times[0,1]\times(0,1)$, essentially \textit{``a collection of stacked-up ROC spaces, with the z-coordinate corresponding to the proportion of the positive class''} \citep{brzezinski_visual-based_2018}.
\cite{flach_geometry_2003} used the term ``isometric" to refer to performance metric contours (or \textit{isolines}) as distinct from  its other meaning---isometric \textit{projection}---in which measurements in different directions represent equal distances. \citeauthor{flach_geometry_2003}'s stacked ROC-space representation involves orthogonal rescaling along each dimension to pack the space of possible confusion matrices into a cube.

The second approach uses projections that preserve the \textit{distances} between different confusion matrices. To explain the notion of distance here, consider the confusion matrix in Table~\ref{tab:confusion-matrix}. By adding 1 to an element and subtracting 1 from another, we can construct 12 different adjacent confusion matrices, e.g.,
\begin{equation*}
    \begin{bmatrix}
    a + 1 & b - 1\\
    c     & d  
    \end{bmatrix},
    \begin{bmatrix}
    a + 1 & b  \\
    c - 1    & d  
    \end{bmatrix},
    \begin{bmatrix}
    a + 1 & b  \\
    c     & d -1   
    \end{bmatrix},
    \dots
    ,
    \begin{bmatrix}
    a    & b  \\
    c -1 & d +1   
    \end{bmatrix}
\end{equation*}
equidistant from the matrix in Table~\ref{tab:confusion-matrix}. The barycentric projection used by \cite{brzezinski_visual-based_2018} ensures these adjacent confusion matrices are equidistant in three dimensional projection.
Rather than producing a cubic stack of square ROC spaces, this projection maps confusion matrices to points that are tetrahedrally packed---different class imbalance ratios yield rectangular cross-sections of this tetrahedron. This is best \href{https://dabrze.shinyapps.io/Tetrahedron}{understood interactively}.

\citet{chicco_benefits_2021} also use a tetrahedral projection (though not projection to a regular tetrahedron) to visualise confusion matrices and it was this work that motivated us to pursue an approach similar to \citeauthor{brzezinski_visual-based_2018} The novel interactive and animated visualisations we propose blend aspects of \citeauthor{flach_geometry_2003}'s contour-based visualisation of performance metric values and enable us to understand the behaviour of these metrics within and beyond ROC space.

\subsection{Studies that consider the uncertainty inherent in empirical performance metric estimates}

In comparison to the volume of literature comparing, contrasting or characterising performance metrics, relatively few authors have considered the issue of \textit{uncertainty} in these metrics. Classifier performance is estimated using finite amounts of test data; the more data we use, and the more representative that data is, the more certain we are about the classifier's performance on new data. With small amounts of test data, uncertainty is high and we must temper our evaluations of classifier performance with that knowledge. Also, it is important to consider the amounts of test data we have \textit{for each class of interest}; while the total amount of test data can be large, a specific class can be rare. When classes are rare, our estimates of a classifier's ability to correctly detect that class become less certain.

Confusion matrices summarise the performance of binary classifiers by counting $\TP$, the number   of ($\Pos$) positive examples, and $\TN$, the number   of ($\Neg$) negative examples that have been successfully classified. Bayesian statistics provides an elegant framework for incorporating prior belief into modeling the predictive distribution of these counts (i.e., the distribution of counts we may expect to see in future trials) through the beta-binomial model \citep{murphy_machine_2012, navarro_introduction_2010, agresti_categorical_2013}.
This approach is used by \cite{totsch_classifier_2021} who further refine Caelen's [\citeyear{caelen_bayesian_2017}] Dirichlet-multinomial model of confusion matrices.

Both \citeauthor{totsch_classifier_2021} and \citeauthor{caelen_bayesian_2017} use \textit{simulation} to draw samples from the posterior distributions of different performance metrics. However, the precise discrete probability mass functions of these metrics can be calculated directly (and quickly) and we do this to allow performance metric uncertainty to be visualised interactively. Our novel approach reveals structure and regularity in the discrete values that performance metrics can take, as well as in their distributions.

This paper is motivated by a desire to better understand the uncertainty inherent in confusion matrix performance metrics and to develop methods that can be used in \textit{specific} scenarios (rather than averaging performance metric behaviour over a range of scenarios). Our hope is that this will enable practitioners to better understand and meaningfully assess the strengths and limitations of classifier performance evaluation with specific evaluation datasets of fixed size and class imbalance. 

Having reviewed prior relevant research, we now present our approach to visualising the distributions of these metrics so that practitioners can put empirical classifier performance statistics into perspective. 

\begin{SCfigure}
  \caption{3D projections of binary confusion matrices of size 100. Each point corresponds to a unique confusion matrix and is coloured by the value of that matrix's Matthews Correlation Coefficient (MCC). For reference, we label the four extreme points corresponding to all True Positives, (TP=100), all False Negatives (FN=100), etc., and connect those vertices to give an impression of the regular tetrahedral lattice (i.e., the 3-simplex) of the projected points. In total, there are $\binom{100+4-1}{4-1}=176\,851$ different binary confusion matrices of size 100. Rather than show all these, we have taken three slices through the lattice: from back to front, the rectangular lattices of points correspond to confusion matrices where $\Pos = 20, 50, 90$, respectively. We have arranged this projection so that axes of these slices are oriented like those of conventional ROC curves: $(\mathrm{TN},\TP)$ are maximum at the top-left and minimum at the bottom-right; the diagonals from bottom-left to top-right correspond to confusion matrices from random classification. Six points are labeled with their corresponding confusion matrices and coloured by their MCC values. This graphic is a composite of screenshots from our \href{https://bit.ly/see-confusion-simplex}{our interactive visualisation} (see Appendix~\ref{app:simplex}).}
  \includegraphics[width=0.6\textwidth]{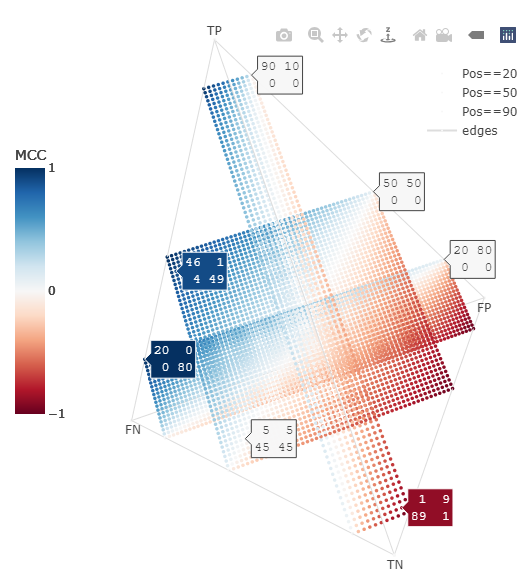}
  \label{fig:confusion-lattice}
\end{SCfigure}

\section{The confusion simplex: representing binary confusion matrices in a 3D lattice of points}

When the four elements of binary confusion matrices sum to a fixed total, these matrices have only three degrees of freedom and can therefore be represented (and visualised) in three dimensions. \citet{chicco_benefits_2021} do this by dividing each element of the confusion matrix by its total and using three of these ratios (specifically $\TP/\Tot$, $\TN/\Tot$ and $\FP/\Tot$) to project the confusion matrix into a \textit{confusion tetrahedron} with vertices $A=(1,0,0)$, $B=(0,1,0)$, $C=(0,0,1)$ and $O=(0,0,0)$ (see \citealp[Figure 1]{chicco_benefits_2021}). If we write the elements of an arbitrary confusion matrix as a row vector $[a, b, c, d]$, then this transformation to the coordinates  $[x, y, z]$ can be written as the matrix product
\begin{align}
\frac{1}{\Tot}
\begin{bmatrix}
a & b & c & d
\end{bmatrix}
\begin{bmatrix}
1 & 0 & 0 \\
0 & 0 & 1 \\
0 & 1 & 0 \\
0 & 0 & 0 
\end{bmatrix}
=
\begin{bmatrix}
x & y & z
\end{bmatrix}.
\label{eq:tetrahedral-transform}
\end{align}
The rows of the transformation matrix can be permuted to provide equivalent projections. Imagine we have the four extreme confusion matrices with a given total, i.e., the matrices where $\TP=\Tot$, $\FP=\Tot$, \textit{etc.}. Equation~\eqref{eq:tetrahedral-transform} maps these matrices to the vertices
\begin{align}
\frac{1}{\Tot}
\begin{bmatrix}
 \Tot &  0  &  0  &  0  \\
 0  & \Tot &  0  &  0  \\
 0  &  0  & \Tot &  0  \\
 0  &  0  &  0  & \Tot 
\end{bmatrix}
\begin{bmatrix}
1 & 0 & 0 \\
0 & 0 & 1 \\
0 & 1 & 0 \\
0 & 0 & 0 
\end{bmatrix}
=
\begin{bmatrix}
1 & 0 & 0 \\
0 & 0 & 1 \\
0 & 1 & 0 \\
0 & 0 & 0
\end{bmatrix}%
\begin{array}{c}
A\\ 
C\\ 
B\\ 
O
\end{array}
\label{eq:confusion-tetrahedon-vertices}
\end{align}
as shown in \cite[Figure 1]{chicco_benefits_2021}. Note that
\begin{itemize}
    \item This is not an isometric projection: the Euclidean distance between points $A, B$ and $C$ is $\sqrt{2}$, while the distance from each of those points to the origin $O$ is $1$.
    \item More importantly, this projection carries only \textit{relative} information about confusion matrices by dividing the counts of true and false positives and negatives by the total. This projection omits important information about how many times we have observed a classifier make predictions about positive and negative examples, and thus, how certain we can be about its predictive performance.
    \item This projection can give the impression of confusion matrices being mapped into a continuous representation rather than a discrete one. However, empirical confusion matrices are discretely valued and belong to the space of four-dimensional natural numbers, $\mathbb{N}^4$.
\end{itemize}
 
With these issues in mind, we propose an isometric projection that \textit{preserves information} about the underlying counts and affords an injective mapping from each possible confusion matrix in $\mathbb{N}^4$ to the space of rationals $\mathbb{Q}^3$:
\begin{align}
\begin{bmatrix}
a & b & c & d
\end{bmatrix}
\begin{bmatrix}
1 & 0 & 0 \\
0 & 1 & 0 \\
0 & 0 & 1 \\
-\tfrac{1}{3} & -\tfrac{1}{3} & -\tfrac{1}{3} 
\end{bmatrix}
=
\begin{bmatrix}
x & y & z
\end{bmatrix}.
\label{eq:simplex-transform}
\end{align}

We refer to the projected points as a \textit{confusion simplex}, to distinguish it from the confusion tetrahedron of \cite{chicco_benefits_2021}. Using a  logic similar to Equation~\eqref{eq:confusion-tetrahedon-vertices}, the vertices of a confusion simplex for matrices of size $N$ are 
\begin{align}
\begin{bmatrix}
\Tot &  0  &  0  &  0  \\
 0  & \Tot &  0  &  0  \\
 0  &  0  & \Tot &  0  \\
 0  &  0  &  0  & \Tot 
\end{bmatrix}
\begin{bmatrix}
1 & 0 & 0 \\
0 & 1 & 0 \\
0 & 0 & 1 \\
-\tfrac{1}{3} & -\tfrac{1}{3} & -\tfrac{1}{3} 
\end{bmatrix}
=
\begin{bmatrix}
\Tot & 0 & 0 \\
0 & \Tot & 1 \\
0 & 0 & \Tot \\
-\tfrac{\Tot}{3} & -\tfrac{\Tot}{3} & -\tfrac{\Tot}{3}
\end{bmatrix}%
\begin{array}{c}
\mathsf{TP} \\ 
\mathsf{FP} \\ 
\mathsf{FN} \\ 
\mathsf{TN}
\end{array}
\end{align}
as shown in Figure~\ref{fig:confusion-lattice} which demonstrates the isometric projection of Eq.~\eqref{eq:simplex-transform} for some of the possible binary confusion matrices of size 100.  Note that
\begin{itemize}
    \item This is an isometric projection: the Euclidean distance between each pair of vertices \textsf{TP}, \textsf{FP}, \textsf{FN} and \textsf{TN} is $N\sqrt{2}$ (e.g., $N=100$ in Figure~\ref{fig:confusion-lattice}).
    \item This projection preserves information about the counts of confusion matrices and, hence, how certain we can be about the performance of the classifier that generated them. 
    \item This projection gives an injective mapping of each possible confusion matrix (in $\mathbb{N}^4$) to a unique  element of the space of rationals $\mathbb{Q}^3$,
\end{itemize}

As we were drafting this document, we learned that a similar isometric projection had been previously proposed by \cite{brzezinski_visual-based_2018}, building on the work of \cite{susmaga_visualization_2015,susmaga_can_2015} to visualise interestingness measures and Bayesian confirmation methods. These works 
are founded on the concepts of barycentric coordinate systems (see, e.g., \cite{floater_general_2006}). \citeauthor{brzezinski_visual-based_2018} give a thorough and detailed analysis of 10 properties of 22 classifier performance metrics in this barycentric projection, as well as  \href{https://dabrze.shinyapps.io/Tetrahedron}{an interactive visualisation} providing valuable insights into  classifier evaluation and confusion matrices.

The barycentric projection used by \citeauthor{brzezinski_visual-based_2018} is described in  \cite[p.320]{susmaga_visualization_2015} and is equivalent to
\begin{align}
\frac{1}{\Tot}
\begin{bmatrix}
a & b & c & d
\end{bmatrix}
\begin{bmatrix}
+1 & +1 & +1 \\
-1 & +1 & -1 \\
-1 & -1 & +1 \\
+1 & -1 & -1 
\end{bmatrix}
=
\begin{bmatrix}
x & y & z
\end{bmatrix}.
\label{eq:barycentric-transform}
\end{align}
As shown in \citet[Figure 1]{susmaga_visualization_2015}, this projection maps the extreme confusion matrices with a given total (i.e., the matrices where $\TP=\Tot$, $\FP=\Tot$, \textit{etc.}) to four of the corners of the cube $[-1,+1]^3$.
Note that, by dividing by the total ($\Tot$), this projection does not map each unique confusion matrix to a unique point. Like \citeauthor{chicco_advantages_2020}'s projection (Equation~\eqref{eq:tetrahedral-transform}), \citeauthor{brzezinski_visual-based_2018}'s projection (Equation~\eqref{eq:barycentric-transform}) carries only \textit{relative} information about confusion matrices. In situations where  information about the actual counts of a confusion matrix matter (as they do in modeling uncertainty), Equation~\eqref{eq:barycentric-transform} could be modified by removing the $1/\Tot$ scale factor, thereby providing a rotated, translated version of the projection we propose in Equation~\eqref{eq:simplex-transform}.

\begin{figure}[t]
    \centering
    \includegraphics[width=\textwidth]{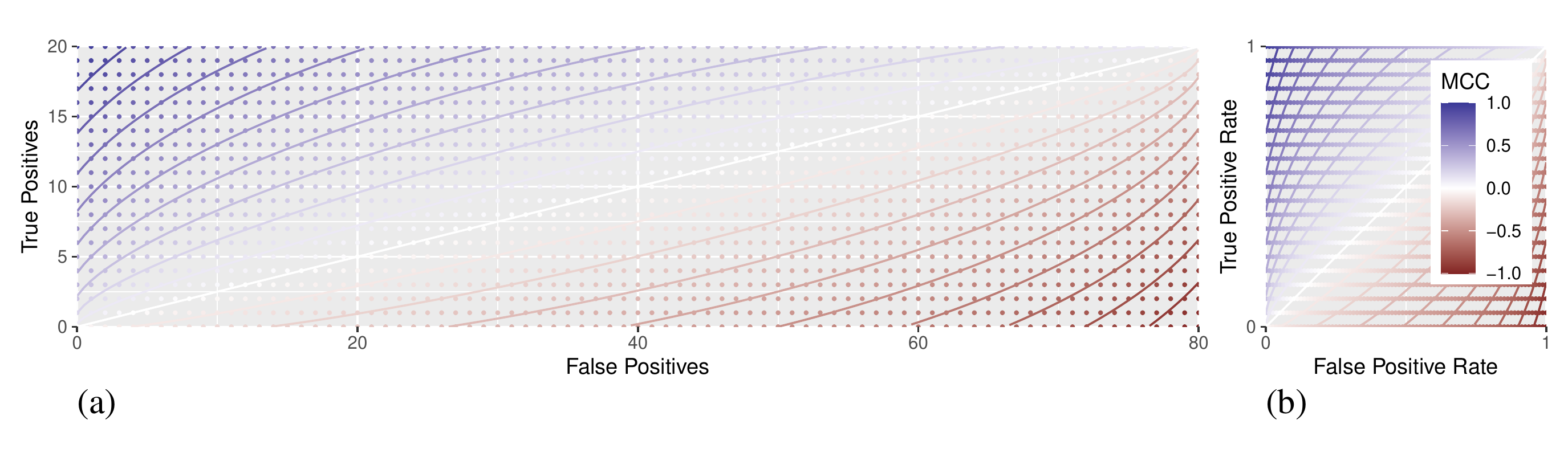}
    \caption{(a) Orthographic projection of the slice  of points from the confusion simplex of Figure~\ref{fig:confusion-lattice} where $\Pos=20$ and $\Neg=80$, coloured by the value of the Matthews Correlation Coefficient (MCC) (Section~\ref{sec:MCC}). The continuous lines indicate the contours of MCC, ranging from $-0.9, -0.8, \dots, 0.9$. Note that while MCC (Equation~\eqref{eq:MCC}) can be calculated for continuous arguments, empirical confusion matrices give rise to a finite set of $(\Pos+1)\times(\Neg+1)$ arguments, corresponding to the points in this 2D lattice.\\
    (b) ROC curves plot a classifier's true positive \textit{rate} against its false positive \textit{rate} in the space of rational numbers from $[0,1]\times[0,1]$. This is equivalent to re-scaling the $x$-axis of (a) by a factor of $1/\Neg$ and the $y$-axis by $1/\Pos$. Again, the contours of the MCC performance metric are defined continuously, but empirical confusion matrices can only take on values at the discrete points in this plot which have $\Neg+1=81$ possible $x$-values ($0, 1/\Neg, 2/\Neg, \dots, 1$) and $\Pos+1=21$ possible $y$-values ($0, 1/\Pos, 2/\Pos, \dots, 1$).}
    \label{fig:slice}
\end{figure}

\section{Going to 2D ROC space by slicing the 3D confusion simplex}

Often, different classifiers are evaluated and compared by presenting them with a fixed set of $\Tot$ examples, of which some ($\Pos$) are positive, and the remainder ($\Neg$) are negative. Interest centres on how many actual positives are correctly identified (true positives: $\TP$) and how many actual negatives are correctly identified (true negatives: $\TN$). Thus, the performance of  classifiers presented with $\Tot = \Pos + \Neg$ examples can be represented in two dimensions using a regular lattice of $(p+1)\times(n+1)$ points.

Figure~\ref{fig:confusion-lattice} shows perspective views of three different 2D slices (rectangular lattices) of points in a confusion simplex: a tall, skinny lattice (where $\Pos=90$, $\Neg=10$); a square lattice where classes are \textit{balanced} ($\Pos=50$, $\Neg=50$); and a short, wide lattice ($\Pos=20$, $\Neg=80$). These slices can be shown in rectangular 2D orthographic projections (e.g., Figure~\ref{fig:slice}(a), \citet[Figure 4]{brzezinski_dynamics_2020}). When the axes of these projections are scaled so that they plot the true positive rate (Equation~\ref{eq:TPR}) against the false positive rate (Equation~\ref{eq:FPR}) of each point, we see these points in the space of the receiver operating characteristic (ROC) curve (e.g., Figure~\ref{fig:slice}).

ROC curves describe the empirical performance of different classifiers and provide a compact and interpretable summary of true and false positive rates. However, by using these rates, ROC curves omit information about the underlying numbers of positive and negative examples used to obtain those empirical performances. This missing information is important because the actual numbers of positive and negative cases in a confusion matrix indicate how certain we are about a classifiers' performance: the more positive and negative cases the classifier has seen, the more certain we are about its performance: true and false positive rates alone do not carry this information.

The aspect ratio of the 2D rectangular sections of the confusion simplex (Figure~\ref{fig:slice}(a)) or barycentric tetrahedron clearly shows this balance of positive and negative examples. However, when classes are highly imbalanced, this approach becomes challenging to print, view and inspect. To deal with this, we suggest faintly plotting all possible points in the ROC space for reference, as demonstrated in Figure~\ref{fig:slice}(b), an approach which is compact, easy to inspect and which clearly shows when there are small numbers of positive or negative examples. 

\begin{figure}[t]
\begin{subfigure}[b]{0.30\textwidth}
    \centering
    \includegraphics[width=\textwidth]{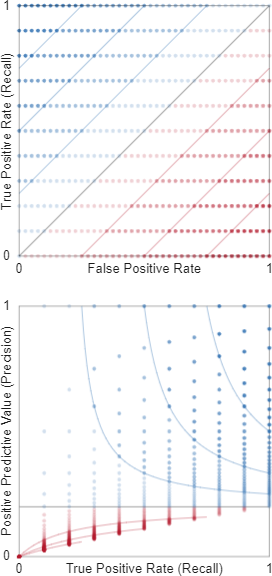}
    \caption{All possible ROC and Precision-Recall points with 10 positive and 40 negative cases (Prev = 0.2).}
    \label{fig:ROC.PRC.10}
\end{subfigure}
\hfill
\begin{subfigure}[b]{0.30\textwidth}
    \centering
    \includegraphics[width=\textwidth]{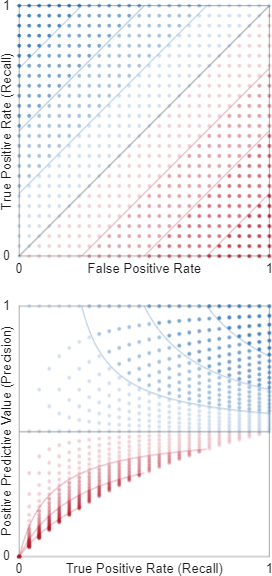}
    \caption{All possible ROC and Precision-Recall points with 25 positive and 25 negative cases (Prev = 0.5).}
    \label{fig:ROC.PRC.25}
\end{subfigure}
\hfill
\begin{subfigure}[b]{0.30\textwidth}
    \centering
    \includegraphics[width=\textwidth]{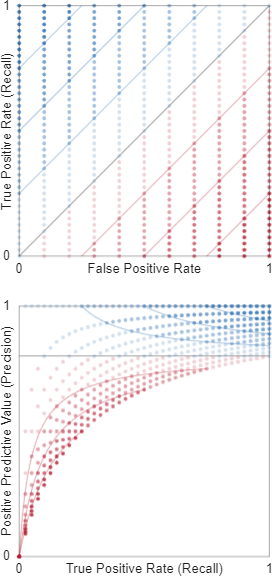}
    \caption{All possible ROC and Precision-Recall points with 40 positive and 10 negative cases (Prev = 0.8).}
    \label{fig:ROC.PRC.40}
\end{subfigure}
\caption{Using reference points in the background of ROC and Precision-Recall plots can indicate the number of positive and negative examples used in a confusion matrix while maintaining a 1:1 plot aspect ratio. The points on these plots correspond to confusion matrices of size 50. We have used colours and reference contours corresponding to Balanced Accuracy (see Section~\ref{sec:BA}) to show the mapping between ROC and Precision-Recall plots, but when plotting actual data, these reference points could be made faint and unobtrusive. These screenshots are taken from \href{http://bit.ly/see-ROC-reference-points}{our interactive visualisations} (see Appendix~\ref{app:ROC}) .}
\label{fig:ROC.PRC}
\end{figure}

As Figure~\ref{fig:ROC.PRC} shows, this pointillist approach can be used with precision-recall plots \citep{davis_relationship_2006} which map $(\FPR,\TPR)$ points in ROC space to $(\TPR,\PPV)$ according to
\begin{align}
    (x,y) \mapsto 
    \left(y, \left(1+\frac{\Neg}{\Pos}\cdot \frac{x}{y}\right)^{-1}\right),
\end{align}
a mapping which clearly depends on class balance through the factor $\tfrac{\Neg}{\Pos}$. While \cite{saito_precision-recall_2015} regard these precision-recall plots as more informative than ROC curves because their achievable shape depends on prevalence, we find them hard to interpret without reference points. We provide \href{http://bit.ly/see-ROC-reference-points}{an interactive visualisation at to show how reference points in ROC and Precision-Recall plots relate} (see Appendix~\ref{app:ROC}).

Having discussed the  visualisation of the discrete lattice of points achievable in ROC and precision-recall space, we next show how to visualise the (continuous) performance metrics of confusion matrices in ROC space.

\section{The geometry of performance metrics in ROC space, and beyond}

Many performance metrics have been proposed as single number summaries of binary confusion matrices (Appendix~\ref{app:metrics}): sometimes the same metric goes by different names (e.g., the Yule $\phi$ coefficient and the Matthews Correlation Coefficient; Youden's $J$ statistic and Bookmaker Informedness); some of these metrics are algebraically related (e.g., Balanced Accuracy and Bookmaker Informedness). Different metrics can have different ranges (e.g., $[0,1], [-1,1], [-\infty,\infty]$) and require further transformation to be compared. Some metrics depend directly on prevalence (a.k.a. class imbalance or skew). There are many metrics to make sense of.

While it is straightforward to see that many metrics achieve their maximum for a confusion matrix where there are only correct classifications, and their minimum when there are no correct classifications, it is less obvious how these metrics behave when the four entries of the confusion matrix are not zero, which is typically of interest when people want to rank or compare the performance of different classifiers.

One approach to gain a better understanding of these metrics is to plot their \textit{contours}, the isolines along which a metric takes a particular value $k$. We have used this to illustrate the curved contours of the Matthews Correlation Coefficient above (Figure~\ref{fig:slice}) and in Appendix~\ref{sec:prevalence-dependent-contours} (Figures~\ref{fig:MCC.contours} and~\ref{fig:MCCbeyond}). \cite{flach_geometry_2003} used this technique to present the geometry of several metrics in the $[0,1]\times[0,1]$ region of ROC space . Other authors have drawn heat-maps or inferred contours from samples of metric values in ROC space.

We go further by deriving algebraic expressions for the exact contours of many popular metrics---both prevalence-dependent (Appendix~\ref{sec:prevalence-dependent-contours}) and prevalence-independent (Appendix~\ref{sec:prevalence-independent-contours})---and provide  \href{http://bit.ly/see-confusion-metrics}{interactive visualisations of these contours} (see Appendix~\ref{app:contours}). We also present animations that show how the contours of different metrics change with prevalence for
\href{https://bit.ly/see-animated-accuracy}{Accuracy}, 
\href{https://bit.ly/see-animated-BA}{Balanced Accuracy}, 
\href{https://bit.ly/see-animated-F1}{$F_1$ score} and
\href{https://bit.ly/see-animated-MCC}{Matthews Correlation Coefficient} (see Appendix~\ref{app:animated-contours}). These animations rely on exact algebraic expressions of contours---they are not estimated from samples of performance metric values (as seen, for example, in \cite{luque_impact_2019}). While sample-based performance metric contour estimates approach the true contours as the sample size approaches infinity, the advantage of having the algebraic expressions in Appendices~\ref{sec:prevalence-dependent-contours} and~\ref{sec:prevalence-independent-contours} is that they show the exact performance metric contours for any size of confusion matrix.

The algebraic expressions of these contours also allow us to appreciate the geometry of performance metrics \textit{beyond} ROC space, as \cite{flach_geometry_2003} had shown for $\fOne$. For example, MCC can be understood as a set of elliptical contours, whose eccentricity depends on class imbalance (Appendix~\ref{sec:MCC}). This is valuable because these contours show how performance metrics vary---including their symmetries \citep{brzezinski_visual-based_2018,luque_impact_2019}---in a way that our visual system can readily apprehend. To complete the picture, we now need to understand how confusion matrices are likely to vary\dots

\section{Modeling uncertainty in confusion matrices}
\label{sec:uncertainty}

Using the notation in Table~\ref{tab:confusion-matrix}, suppose a classifier is presented with $\Pos_1$ positive examples to classify and that it gets $a_1$ of these correct (true positives) and the remainder $c_1=\Pos_1-a_1$ incorrect (false negatives). On the basis of these observations, what do we believe is the probability ($\theta_a$) that this classifier correctly identifies positive examples?

A frequentist approach would estimate that $\theta_a$ is the empirical true positive rate of the classifier
\[
\theta_a = \frac{a_1}{p_1} = \TPR
\]
and that the distribution of $a$ future correct classifications with $\Pos$ new examples is
\begin{align}
    a | \Pos, \theta_a\ \sim\ \mathrm{Binomial}(\Pos, \theta_a)
    \label{eq:binomial-dist-TP}
\end{align}
so that
\begin{align}
P(a | \Pos, \theta_a) = \binom{a}{\Pos} \theta_a^{a} (1 - \theta_a)^{p - a}.
    \label{eq:binomial-pmf-TP}
\end{align}

A Bayesian approach \citep{navarro_introduction_2010,murphy_machine_2012,agresti_categorical_2013} allows us to express our prior uncertainty about a classifier's true positive rate---this is particularly important when we have small amounts of data. We can assign a prior distribution to $\theta_a$, and it is mathematically convenient to do that with a beta distribution:
\[
\theta_a|u,v\ \sim\ \mathrm{Beta}(u, v).
\]
Because this beta prior is conjugate to the binomial distribution of successes, after observing $a_1$ true positive (and $c_1$ false negative) classifications of $p_1$ examples, the posterior distribution of $\theta_a$ remains a beta distribution:
\[
\theta_a|a_1, p_1, u,v\ \sim\ \mathrm{Beta}(u+ a_1, v+c_1)
\]
and the posterior predictive distribution of seeing $a$ correct classifications when $p$ further examples are presented  is
\begin{align}
\mathrm{P}(a \ | \ p, a_1, c_1, u, v)
&=
\binom{p}{a}
\frac{\mathrm{Beta}(u+a_1+a,v + c_1 +c)}
{\mathrm{Beta}(u+a_1,v + c_1 )}
\label{eq:beta-binomial-pmf}
\end{align}
We write that $a$ is distributed according to a \textit{beta-binomial} distribution:
\begin{align}
a \ | \ p, a_1, c_1, u, v\ &\sim\ 
\mathrm{BB}(a, p, u + a_1,v + c_1)
\label{eq:beta-binomial-dist}
\end{align}

Equations~\eqref{eq:binomial-dist-TP} and~\eqref{eq:binomial-pmf-TP} give us the basis of a \textit{binomial} model for uncertainty in confusion matrices and Equations~\eqref{eq:beta-binomial-pmf} and~\eqref{eq:beta-binomial-dist} give us the basis of a \textit{beta-binomial} model. These equations deal with the distribution of the number of true positives returned by classifier presented with positive examples. Using the same logic, we can express a binomial model for the number of true negatives ($d$) returned by classifier presented with $n$ negative examples when the probability of the classifier correctly identifying these negative examples is $\theta_d$:
\begin{align}
    d | \Neg, \theta_d\ \sim\ \mathrm{Binomial}(\Neg, \theta_d)
    \label{eq:binomial-dist-TN}
\end{align}
so that the probability of seeing $d$ true negatives in $n$ negative examples is
\begin{align}
P(d | \Neg, \theta_d) = \binom{d}{\Neg} \theta_d^{d} (1 - \theta_d)^{\Neg - d}.
    \label{eq:binomial-pmf-TN}
\end{align}
where $\theta_d$ is estimated as the empirical true negative rate of the classifier
\[
\theta_d = \frac{d_1}{n_1} = \TNR.
\]

The beta-binomial model of true negatives is thus
\begin{align}
d \ | \ \Neg, d_1, b_1, u, v\ &\sim\ 
\mathrm{BB}(d, \Neg, u + d_1,v + b_1)
\label{eq:beta-binomial-dist-TN}
\end{align}
where $d_1$ is the observed number of true negatives and $b_1$ the observed number of false positives.
The posterior predictive distribution of seeing $d$ correct classifications when a further $\Neg$ negative examples are presented to the classifier is
\begin{align}
\mathrm{P}(d \ | \ \Neg, d_1, b_1, u, v)
&=
\binom{n}{d}
\frac{\mathrm{Beta}(u+d_1+d,v + b_1 +b)}
{\mathrm{Beta}(u+d_1,v + b_1 )}.
\label{eq:beta-binomial-pmf-TN}
\end{align}
Using beta-binomial models for true positives and true negatives demands that we declare our prior uncertainty about $\theta_a$ and $\theta_d$. For demonstration, and in the absence of other relevant information, we choose the uninformative uniform priors:
\begin{align}
\theta_a\ \sim\ \mathrm{Beta}(1, 1)
\qquad
\theta_a\ \sim\ \mathrm{Beta}(1, 1).
\label{eq:uniform-priors}
\end{align}
A common scenario in classifier development and evaluation is to present a classifier with a test set of $\Pos_1$ positive and $\Neg_1$ negative examples. In this situation, we can assume that the probabilities of correctly classifying a further $a$ positive and $d$ negative examples are statistically independent, so that
\begin{align}
    P(a,d|a_1, p_1, d_1, n_1) = P(a|a_1, p_1) \cdot P(d|, d_1, n_1)
\end{align}
for the binomial model. The joint distribution of true positives and negatives under a beta-binomial model can be factored in a similar way, paving the way for confusion matrix uncertainty to be visualised.

We note that there are potential situations where this independence assumption may not hold, for example in cytopathology analysis, if apparently normal cells are extracted from a sample obtained from a positive patient in addition to abnormal cells \citep{burger_changes_1981}.

\section{Visualising uncertainty Part I: confusion matrix distributions}

So far, the ideas set out in the previous section are the same as those set out by \cite{totsch_classifier_2021} who proceed to use simulation to estimate the posterior distribution of various performance metrics under beta-binomial models of true positives and negatives, building on Caelen's [\citeyear{caelen_bayesian_2017}] Bayesian interpretation of the confusion matrix.

However, as \citeauthor{totsch_classifier_2021} point out, \textit{``the posterior distribution can be derived analytically. There is no need for Markov chain Monte Carlo sampling''}. So, rather than using time-consuming simulation, we have used the probability mass functions of Equations~\eqref{eq:binomial-pmf-TP} and~\eqref{eq:binomial-pmf-TN} to develop \href{http://bit.ly/see-confusion-uncertainty}{an interactive visualisation of the impact of uncertainty on confusion matrices and their performance metrics} in ROC space and in Precision-Recall space (see Appendix~\ref{app:uncertainty}). This visualisation uses Desmos' Graphing Calculator \citep{desmos_inc_desmos_nodate} which we have found to be a flexible, intuitive and very convenient platform for interaction with mathematical expressions, and one which supports reproducible research by making these expressions open and accessible.

\begin{figure}[p]
\begin{subfigure}[b]{0.49\textwidth}
    \centering
    \includegraphics[width=\textwidth]{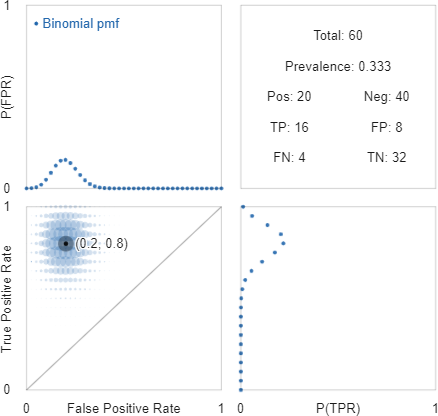}
    \caption{Posterior predictive pmfs under a binomial model.}
    \label{fig:Binomial-pmf}
\end{subfigure}
\hfill
\begin{subfigure}[b]{0.49\textwidth}
    \centering
    \includegraphics[width=\textwidth]{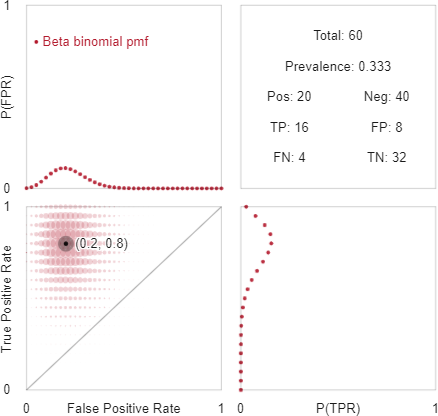}
    \caption{Posterior predictive pmfs under a beta-binomial model}
    \label{fig:Beta-Binomial-pmf}
\end{subfigure}
    \caption{Joint and marginal posterior predictive probability mass functions (pmfs) under different models of uncertainty: blue for binomial; red for beta-binomial. Both subfigures relate to the same confusion matrix (top right panels) and have the same panel layout. These graphics depict the distributions we would expect to see if the classifier that produced that confusion matrix was given a further 20 positive and 40 negative examples to classify. Top left panels show the pmfs of the false positive rate; bottom right panels show the pmfs of the true positive rate; bottom left panels show the joint pmfs of true and false positive rates (i.e., the posterior predictive distribution in ROC space).  The areas of the circles in the joint distribution plots are proportional to the probability masses at the points they are centred on.  These are screenshots from  \href{http://bit.ly/see-confusion-uncertainty}{our interactive visualisation} (see Appendix~\ref{app:uncertainty}).}
    \label{fig:pmfs}
\end{figure}

\begin{figure}[p]
\begin{subfigure}[b]{0.49\textwidth}
    \centering
    \includegraphics[width=\textwidth]{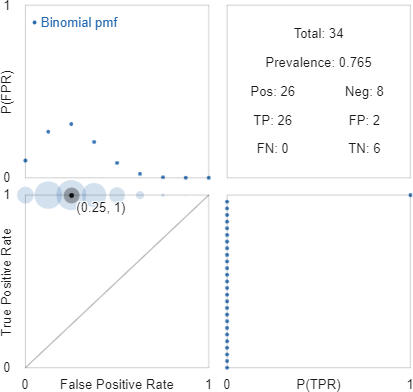}
    \caption{Posterior predictive pmfs under a binomial model.}
    \label{fig:Binomial-pmf-FN0}
\end{subfigure}
\hfill
\begin{subfigure}[b]{0.49\textwidth}
    \centering
    \includegraphics[width=\textwidth]{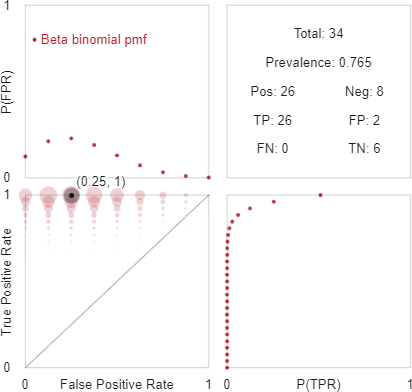}
    \caption{Posterior predictive pmfs under a beta-binomial model}
    \label{fig:Beta-Binomial-pmf-FN0}
\end{subfigure}
    \caption{Joint and marginal posterior predictive probability mass functions (pmfs) of the cocaine purity classifier confusion matrix reported by \cite{rodrigues_analysis_2013} and used by \cite{totsch_classifier_2021} to illustrate confusion matrix uncertainty. These are screenshots from  \href{http://bit.ly/see-confusion-uncertainty}{our interactive visualisation} (see Appendix~\ref{app:uncertainty}).}
    \label{fig:pmfs-FN0}
\end{figure}

Figure~\ref{fig:pmfs} shows two screenshots from our interactive visualisation that demonstrate how binomial (in blue) and beta-binomial (in red) model assumptions convey different levels of uncertainty about a classifier's future predictions.  Unlike the histograms of samples from the posterior predictive distributions of true positive and true negative rates shown by \cite{totsch_classifier_2021}, the visualisations of the probability mass functions in Figure~\ref{fig:pmfs} remind us of the underlying discreteness of ROC space. Note that the pmfs of the beta-binomial model are broader than those of the binomial, reflecting the additional uncertainty in true and false positive rates embodied in this Bayesian model.

To further illustrate the differences between the uncertainties conveyed by these two models, we consider the drug purity data from \cite{rodrigues_analysis_2013} used by \cite{totsch_classifier_2021}. Figure~\ref{fig:pmfs-FN0} visualises the confusion matrix of 26 positive examples and 8 negative examples in which no false negatives were observed. The binomial model of the true positive rate places the entire probability mass at $\TPR=1$ (Figure~\ref{fig:Binomial-pmf-FN0}, bottom right), while the beta-binomial is more moderate, suggesting that a range of true positive values from 0.8--1.0 are plausible (Figure~\ref{fig:Beta-Binomial-pmf-FN0}, bottom right).  \cite{totsch_classifier_2021}[Figure 4] explored uncertainty in this data by simulating $20\,000$ draws from the posterior predictive distribution of the beta-binomial model and using histograms to summarise the true positive and negative rates that were sampled. This work and the foundation that \cite{caelen_bayesian_2017} provided highlight the importance of modeling uncertainty in interpreting confusion matrices. We think that calculating and visualising the exact discrete probability mass functions conveys an even more meaningful and accurate appreciation of that uncertainty.

The more labelled data evaluated by a classifier, the more certain we can be about its true and false positive rates. Figure~\ref{fig:pmfs.by.N} illustrates this by showing the posterior predictive pmfs of confusion matrices of increasing totals but constant true and false positive rates: more data yields more precise estimates.

\begin{figure}[p]
    \centering
    \includegraphics[width=\textwidth]{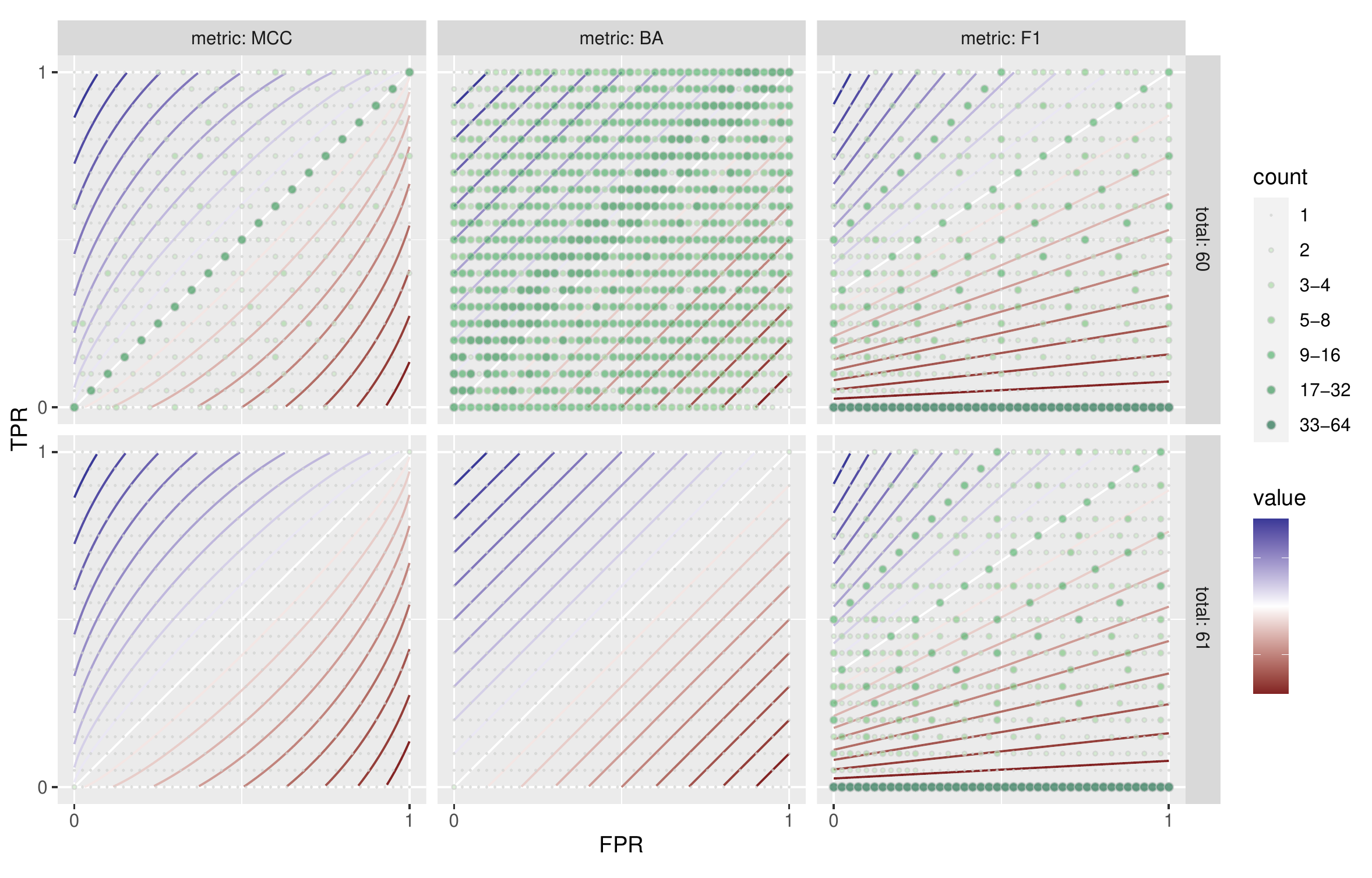}
    \caption{Each panel shows all the possible points in the ROC space of confusion matrices of 20 positive and 40 negative examples (top row) and 20 positive and \textit{41} negative examples (bottom row). Points are coloured by the number of times the performance metric value at that point is observed in the confusion matrices of those totals. Three different performance metrics are presented: MCC (left), BA (middle), $\fOne$ (right). Performance metric contours are shown in the background, coloured by their value. Note that one additional negative example changes the configuration of possible points in ROC space so that each possible MCC and BA value is unique (bottom left and middle); the multiplicity of different $\fOne$ values remains much the same. (bottom right).}
    \label{fig:counts}
    \includegraphics[width=0.7\textwidth]{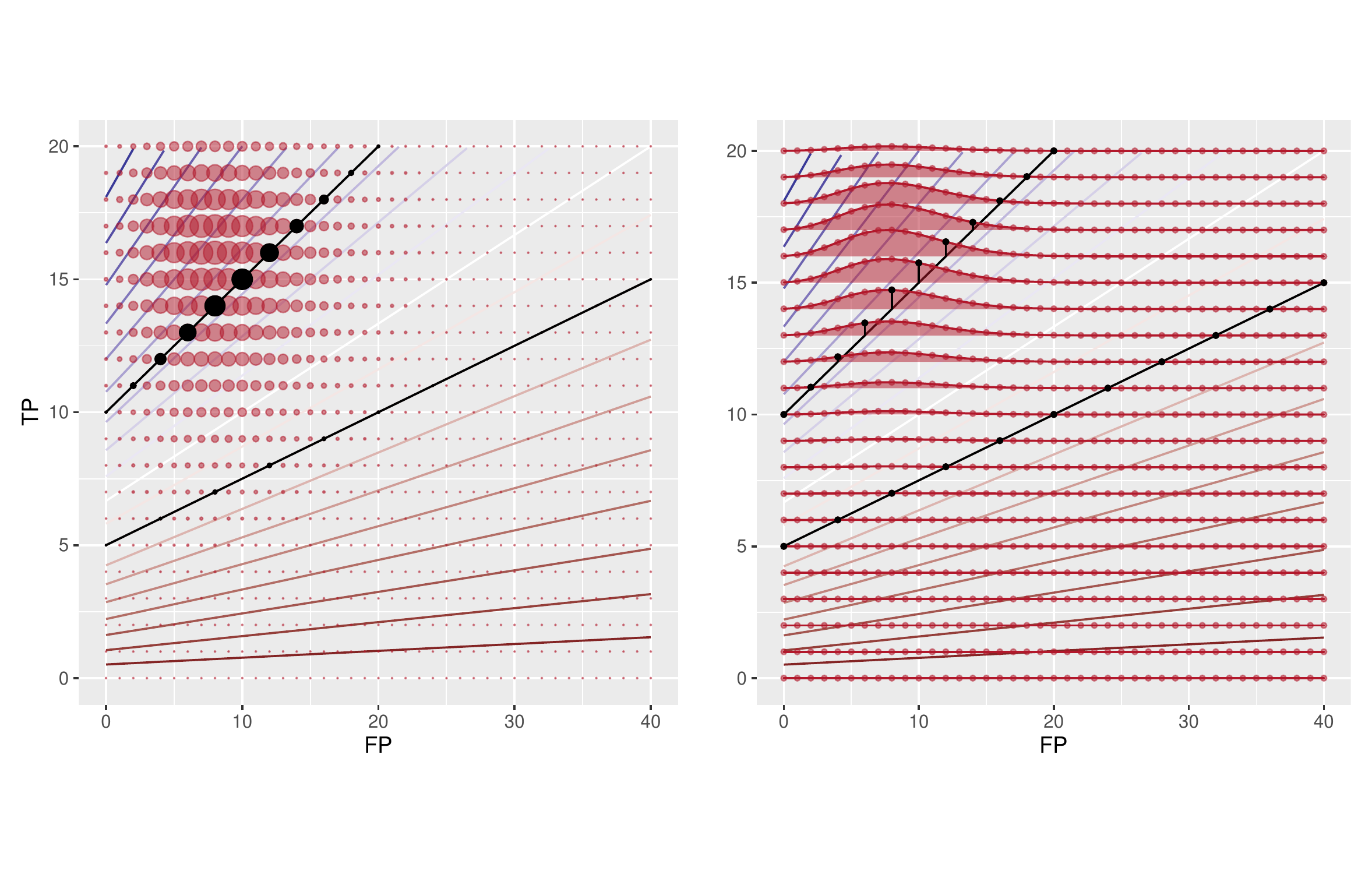}
    \vspace{-0.7cm}
    \caption{Two ways to show confusion matrix pmfs and performance metric contours in ROC space. Both plots show the posterior predictive pmf of confusion matrices under a beta-binomial model of uncertainty for a classifier observed to produce the confusion matrix $\SmallMatrix{16&8\\4&32}$ (the same as in Figure~\ref{fig:pmf.60}(b)). Like Figure~\ref{fig:pmf.60}, the left plot uses circle areas to represent probability mass; the right plot uses ridge lines. In the background are the contours of the $\fOne$ performance metric and in black are the contours $\fOne=\tfrac{4}{10}$ and $\fOne=\tfrac{2}{3}$, along each of which lie 11 points in ROC space.}
    \label{fig:pmfs.F1}
\end{figure}

\section{Visualising uncertainty Part II: performance metric distributions}

Having calculated and visualised the posterior predictive pmf of confusion matrices of size $\Tot=\Pos+\Neg$ after observing a classifier's empirical performance, we can now visualise the distribution of a given performance metric. Let $\mathbf{C}$ represent the confusion matrix random variable whose pmf is $P_\mathbf{C}(\mathbf{c})$, and let $M=\mu(\mathbf{C})$ represent the random variable we get by applying a performance metric function $\mu(\cdot)$ to a confusion matrix. The distribution of that performance metric is
\begin{align*}
    P_M(m) &= P(\mu(\mathbf{C})=m)\\
    &= \sum_{\mathbf{C}:\mu(\mathbf{C})=m} P_\mathbf{C}(\mathbf{c})
\end{align*}
i.e., the probability mass at performance metric value $m$ is sum of the probability masses where $\mu(\mathbf{C})=m$. In other words, we find the pmf of the performance metric by summing the probability masses that lie along each contour of the performance metric in ROC space. 


The geometry of performance metric contours in conjunction with the layout of the $(\Neg+1)\times(\Pos+1)$ possible points in ROC space determines which probability masses are summed together, as illustrated in Figure~\ref{fig:counts}. With confusion matrices of 20 positive and 40 negative examples (top row) certain MCC, BA and $\fOne$ contours intersect multiple points in ROC space, most noticeably the $0$ contours, which intersect 21 points with MCC and BA (along the $(0,0),(1,1)$ diagonal), and 41 points with $\fOne$ (along the $(0,0),(1,0)$ horizontal). One additional negative example (Figure~\ref{fig:counts}, bottom row) removes this confluence of points and contours in ROC space for MCC and BA, but not $\fOne$.

\begin{figure}[p]
    \centering
    \includegraphics[width=\textwidth]{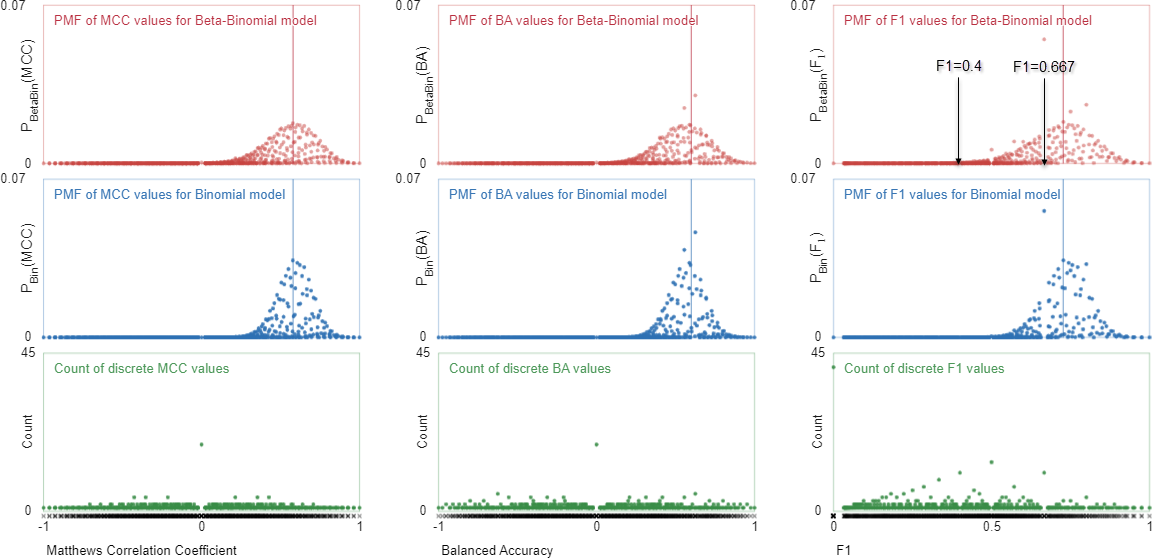}
    \caption{Posterior predictive pmfs for different performance metrics (left to right: MCC, BA, $\fOne$) under beta-binomial (top row, red) and binomial models (middle row, blue) of uncertainty, given an observed confusion matrix of 20 positive and 40 negative examples $\SmallMatrix{16&8\\4&32}$ as used in Figure~\ref{fig:pmfs}. Vertical lines show the MAP performance metric values. The bottom row (green) counts the number of times each value of a particular performance metric is observed: these counts are proportional to the prior pmfs of performance metric values. $\fOne$ values of $\tfrac{4}{10}$ and $\tfrac{2}{3}$ are observed 11 times as shown in Figure~\ref{fig:pmfs.F1}. These are screenshots from  \href{http://bit.ly/see-confusion-uncertainty}{our interactive visualisation} (see Appendix~\ref{app:uncertainty}). The key point to note is that there is potential for substantial variation about the performance metric values of the observed confusion matrix.}
    \label{fig:pmf.60}
\end{figure}

\begin{figure}[p]
    \centering
\includegraphics[width=\textwidth]{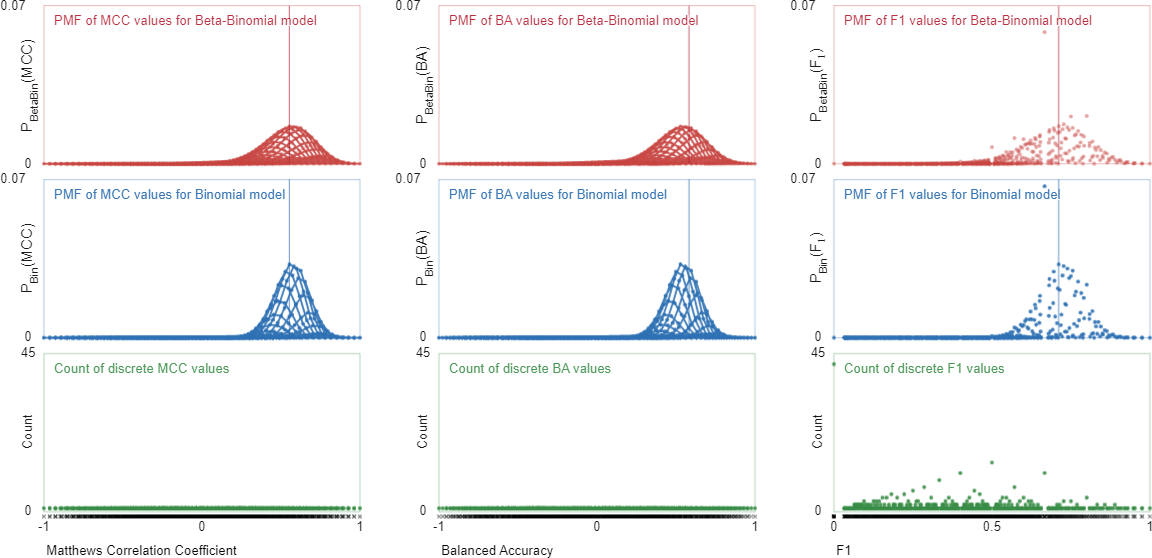}
    \caption{These plots visualise the same quantities as Figure~\ref{fig:pmf.60} but with a slightly different observed confusion matrix of 20 positive and 41 negative examples: $\SmallMatrix{16&9\\4&32}$, i.e., one more false negative. Note the changes in the counts of discrete MCC and BA values in comparison to Figure~\ref{fig:pmf.60}. To emphasise the now smoothly-changing pmf functions for MCC and BA, we use lines to connect points corresponding to the same true positive rate, (i.e., horizontal slices of the joint pmfs in Figure~\ref{fig:pmfs}).}
    \label{fig:pmf.61}
\end{figure}

Using $\fOne$ as an example, Figure~\ref{fig:pmfs.F1} demonstrates how we can visualise the posterior predictive pmfs of confusion matrices in relation to performance metric contours. Plotting these probability masses against performance metric values gives posterior predictive distributions such as shown in the top and middle rows of Figures~\ref{fig:pmf.60} and~\ref{fig:pmf.61}.
The top and middle rows of these Figures show the posterior predictive pmfs of MCC, BA and $\fOne$ values, given observations of $\SmallMatrix{16&8\\4&32}$ (Figure~\ref{fig:pmf.60}) and $\SmallMatrix{16&9\\4&32}$ (Figure~\ref{fig:pmf.61}). The first point to note about these pmfs is their spread around the \textit{maximum a posteriori} (MAP) value of each performance metric (indicated by the vertical line). This spread corresponds to the posterior distributions of true and false positive rates shown in Figure~\ref{fig:pmfs} and serves to put performance comparisons into perspective: MCC, BA and $\fOne$ each show substantial uncertainty about the values they take with the observed confusion matrices. 

Note also the shape of the pmfs in Figure~\ref{fig:pmf.60}: overall they are bell-shaped and look to be comprised of smaller bell-shaped series of points. However, both BA and $\fOne$ have a few points lying above the general bell-curve of points. These correspond to situations where several points in ROC space lie on the same performance metric contour (e.g., $\fOne=\tfrac{4}{10}$ and $\fOne=\tfrac{2}{3}$ as shown in Figure~\ref{fig:pmfs.F1}). The numbers of points that lie on each performance metric contour are shown in green in the bottom row of Figure~\ref{fig:pmf.60}. Adding one extra negative example (as shown in Figure~\ref{fig:pmf.61}) markedly changes the confluence of ROC points and performance metric contours for MCC and BA, less so for $\fOne$.
Now, the MCC and BA pmfs change smoothly, following bell-shaped curves as we sweep across each row (i.e., true positive rate) of the joint pmfs in the bottom left panels of Figure~\ref{fig:pmfs} (a) and (b)---we emphasise this by using lines to connect these probabilities in Figure~\ref{fig:pmf.61}, just like the ridge line plot in Figure~\ref{fig:pmfs.F1}. The posterior pmf of $\fOne$ remains much the same, due to the particular linear relationship between its contours in ROC space and the values of $\Pos$ and $\Neg$ (see Section~\ref{sec:F1}). 

\begin{figure}[t]
    \centering
    \includegraphics[width=\textwidth]{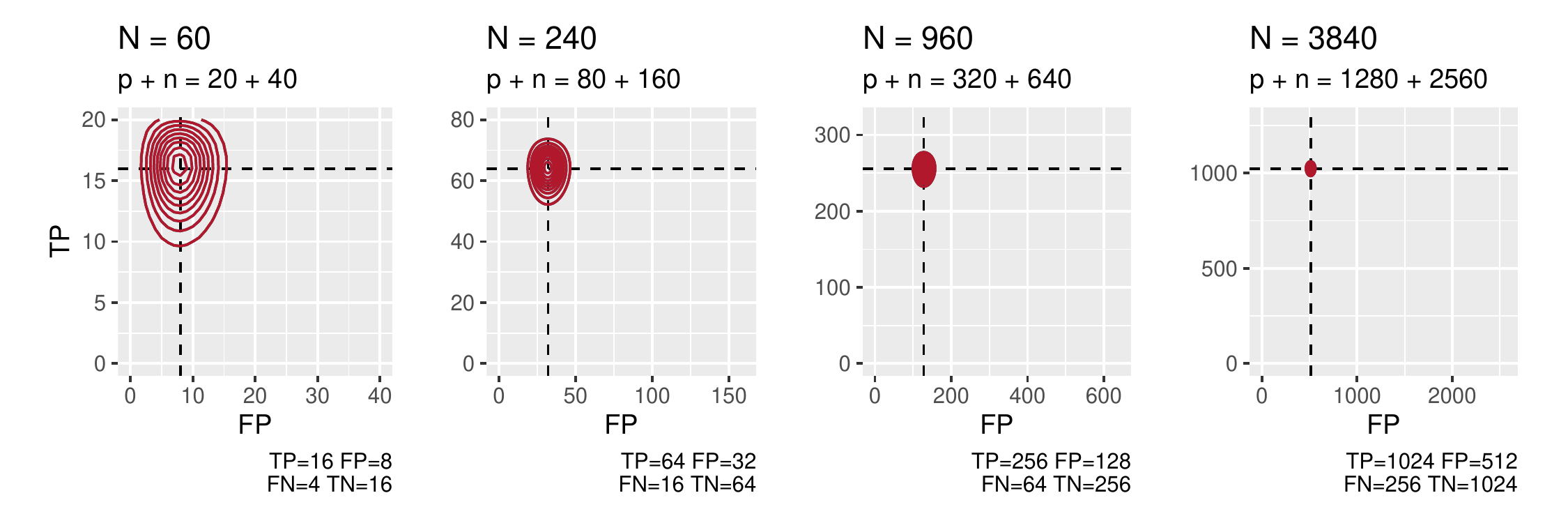}
    \caption{Reducing uncertainty in performance metrics requires more data to increase the precision of the predictive distribution of confusion matrices. These four contour plots show the posterior predictive pmfs (under a beta-binomial model of uncertainty) after observing confusion matrices of increasing size but with the same false and true positive rates (0.2, 0.8). From left to right, the sizes of confusion matrix increase by a factor of 4 and the heights and widths of the contours decrease by a factor of $\tfrac{1}{2}$.}
    \label{fig:pmfs.by.N}
\end{figure}

In summary, we have shown how the discrete nature of confusion matrices can lead to jumps in the pmfs of performance metrics, even though those metrics are smooth continuous functions.
In practical terms, this is a minor issue though. The main point to note from this section is that there can be significant uncertainty in performance metrics,  uncertainty which depends on the amount of data used in evaluating empirical performance, not the performance metrics themselves. To reduce that uncertainty requires more data, and data that is representative of the cases that the classifier will see in production. Roughly speaking, under the binomial and beta-binomial models of uncertainty, the standard deviation of the true and false positive rate pmfs will be proportional to $\tfrac{1}{\sqrt{\Tot}}$, as illustrated in Figure~\ref{fig:pmfs.by.N}.

\section{Discussion} 

In this paper, we aim to draw attention towards the issue of uncertainty in empirical estimates of classifier performance and that performance metrics do not serve to reduce that uncertainty: greater certainty demands greater data. We appreciate that performance evaluation and model selection can involve a host of competing considerations beyond predictive performance, such as model transparency and fairness to different groups affected by model predictions; these are not within the scope of our work here. Nor do we address the common assumption that evaluation data is representative, i.e., that future data are expected to come from the same process as past data and have roughly the same range of values \citep{mcelreath_statistical_2016}. Our goal is to ensure that practitioners can understand, visualise and put into perspective the magnitude and nature of uncertainty in classifier performance estimates so that this can inform more meaningful discussion of the strengths and limitations of specific classification systems. 

\subsection{Incorporating decision costs}

So far, the performance metrics we have discussed summarise the 3 dimensions of variation in binary confusion matrices without regard for the costs of incorrect decisions. One could argue that the best classifier for a given scenario minimises the \textit{costs} (or maximizes the \textit{benefits}) we would expect to see as a result of its future decisions. (Note that we describe the contours of decision costs and benefits in Section~\ref{sec:decision-costs}.)

This is a conceptually appealing way to describe classifier performance with a single number but, in practice, adds further uncertainty into the evaluation process, i.e., uncertainty about the costs or benefits of different decisions. Eliciting and quantifying these costs is difficult and subjective---imagine assessing the costs of different outcomes for a disease diagnosis system from the point of view of a clinician, or from the point of view of a patient. Typically, the vexed problem of decision costs is ignored by using performance metrics that implicitly treat the costs of false positives and negatives as equal. A more explicit approach in reporting performance evaluations---especially to a general audience--- would be to state clearly that ``this evaluation treats false alarms and missed detections as equally costly.''

\subsection{The ``problem'' of class imbalance}

Considerable attention has been given to the ``problem'' of class imbalance (see, e.g., \cite{jeni_facing_2013,mosley_balanced_2013,luque_impact_2019,mullick_appropriateness_2020,lovell_taking_2021}). We use quotes to emphasise that class imbalance is mainly problematic to those of us trying to build automated decision making systems---the real problem of rare but highly adverse cases (e.g., life threatening disease) is that they occur at all; we would not want them to happen more frequently. Also, class imbalance is a necessary consequence of multinomial classification: when there are $C$ possible classes and $\Tot$ examples, at least one class will have equal or fewer than $N/C$ examples.

There are two types of confusion matrix performance metric: \textit{prevalence-independent} (or \textit{balanced}) metrics, whose values depend only on the rates of true positives and negatives (Appendix~\ref{sec:prevalence-independent-contours}); and \textit{prevalence-dependent} metrics whose values depend on these rates and prevalence (i.e., class balance; Appendix~\ref{sec:prevalence-dependent-contours}). (Note that balanced versions of prevalence dependent metrics can be derived---see \cite{luque_impact_2019} and Section~\ref{sec:balanced-FM} for example.) This has led to discussion about which performance metrics are best for scenarios where classes are balanced or imbalanced. Neither balanced nor prevalence-dependent performance metrics reduce the uncertainty arising from finite amounts of evaluation data---only additional data can do that. Figure~\ref{fig:pmfs-FN0} exemplifies scenarios where the problem is not \textit{``which performance metric is best to assess this classifier?''} but rather \textit{``how can we reduce uncertainty about the classifier's detection ability?''}

We note that augmenting examples of rare classes to assist with learning and performance evaluation will distort class prior probabilities, and the trained classifier's outputs will have to be adjusted to provide accurate predictions for real-world class abundance \cite{saerens_adjusting_2002}.

\subsection{Extensions from binary to multinomial classification}

We see two possible ways to extend what we have presented for binary confusion matrices and metrics to multinomial classification, where there are more than two classes to consider.

The first approach is to treat a $C\times C$ multinomial confusion matrix as a set of $C$ binary classification problems, in which each class, in turn, is considered against all others. This \textit{one-versus-all} strategy is common in multinomial classification and has been used to provide compact, informative visualisations of confusion matrices in terms of their prior and posterior odds \citep{lovell_taking_2021}. The binomial and beta-binomial models of uncertainty we have presented here could be applied to each class versus all others, yielding posterior predictive pmfs of various performance metrics. One advantage of this approach would be to reveal the degree of uncertainty associated with each class so that those developing classification systems could consider where best to direct their attention in making improvements.

The second approach would be to change the probabilistic model of confusion matrix distributions from binomial (or beta-binomial) to multinomial (or Dirichlet-multinomial) \citep{murphy_machine_2012}. One of the challenges with this is that with $C>2$ classes, we can no longer visualise the space of $C\times C$ confusion matrices in 3 dimensions.

%
%

\section{Conclusion}

Publications advocating for or against specific performance metrics have motivated us to look closer at the (continuous) geometry of performance metrics and the (discrete) geometry of the space of confusion matrices they are applied to. In doing so, we have gained a clearer understanding of the cause and effect of uncertainty in performance metrics: the primary cause is the uncertainty we have about the confusion matrices that will be produced by a trained classifier; this depends on the numbers of actual positive and negative examples we have observed the classifier determine. Using binomial and (the more conservative) beta-binomial models of uncertainty, we can calculate the exact pmfs of the predictive distribution of confusion matrices, given the classifications we have observed and an initial prior belief in the distribution of a classifier's true and false positive rates.

Using the contours of various performance metrics, we have demonstrated how the posterior predictive pmfs of confusion matrices can be transformed into posterior predictive pmfs of different performance metrics. We have provided a range of static and interactive visualisations for researchers and practitioners to explore uncertainty in confusion matrices and various performance metrics derived from them. These contributions aim to put performance metrics and their uncertainty into perspective, specifically, when observations of positive or negative classes are few, uncertainty in performance matrices can easily eclipse differences in performance between classifiers. Arguments about which classifier performs best, or which performance metric is best, need to take this uncertainty into account, and the visualisations we provide enable researchers to do that.

Our work also provides a useful perspective on performance evaluation where classes are imbalanced---a common scenario in binary decision making and a necessary situation in multinomial classification. Some performance metrics depend on class imbalance; others do not; and ``balanced'' metrics can be created from prevalence-dependent ones. However, the fundamental issue in performance evaluation is not so much the \textit{relative} abundance of different classes as their \textit{absolute} abundance: our estimates of a classifier's ability to correctly detect a class will be highly uncertain when there are few instances that class, regardless of balance. Performance metrics can't address this issue. More data is needed and, until it arrives, we must acknowledge the uncertainty in our findings.

We acknowledge that classifier performance evaluation goes far beyond purely quantitative metrics. It is heartening to see the breadth of issues in performance evaluation, benchmarking and datasets gaining more attention. Still, quantitative metrics will always play a prominent role in considering the strengths and limitations of classification systems. We hope that this visualisation of confusion matrix performance metrics and their uncertainties will inform their use in practice. 

\newpage
\bibliographystyle{unsrtnat}
\bibliography{confusR-PRL}  
\newpage
\appendix

\section{Performance metrics and other definitions}
\label{app:metrics}

Many sums, products, ratios and functions of the four elements of a binary confusion matrix have been defined. Unfortunately, the notation varies between authors as does the row and column layout of the confusion matrix itself. We define performance metrics using the notation and layout of the confusion matrix in Table~\ref{tab:confusion-matrix} in the main paper:

\begin{center}
    \input{table.binary}
\end{center}

\input{eqn.statistics}

\newpage

\section{Performance metric contours that depend on prevalence}
\label{sec:prevalence-dependent-contours}


\subsection{Matthews Correlation Coefficient}
\label{sec:MCC}


\begin{figure}[t]
\begin{subfigure}[b]{0.33\textwidth}
    \centering
    \includegraphics[width=\textwidth]{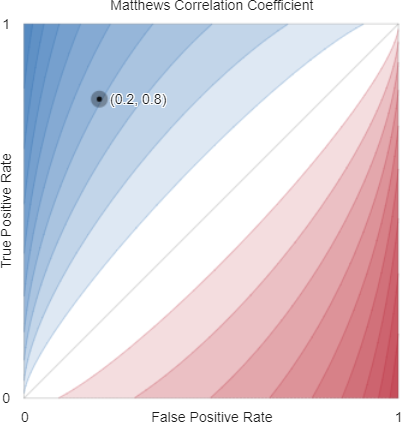}
    \caption{MCC with Prev = 0.1}
    \label{fig:MCC.p10}
\end{subfigure}
\begin{subfigure}[b]{0.33\textwidth}
    \centering
    \includegraphics[width=\textwidth]{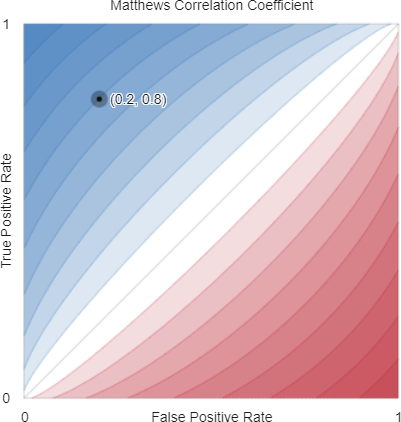}
    \caption{MCC with Prev = 0.5}
    \label{fig:MCC.p50}
\end{subfigure}
\begin{subfigure}[b]{0.33\textwidth}
    \centering
    \includegraphics[width=\textwidth]{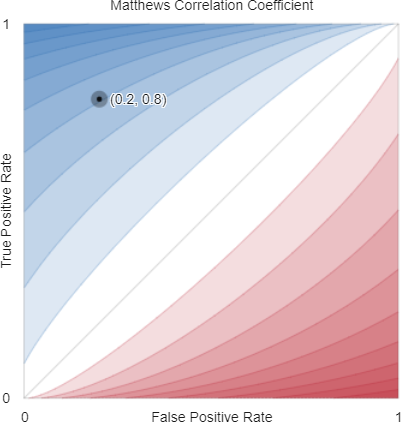}
    \caption{MCC with Prev = 0.9}
    \label{fig:MCC.p90}
\end{subfigure}
\caption{Contours of the Matthews Correlation Coefficient in the ROC space.}
\label{fig:MCC.contours}
\end{figure}

\begin{figure}[t]
\begin{subfigure}[b]{4.5cm}
\centering
\includegraphics[width=4cm]{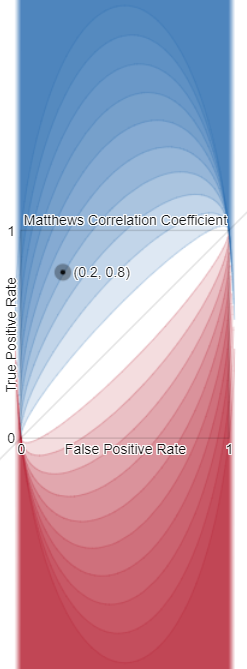}
\caption{MCC with Prev=0.1}
\end{subfigure}
\begin{minipage}[b]{\dimexpr\textwidth-5cm}
\begin{subfigure}[b]{\textwidth}
\centering
\includegraphics[width=5cm]{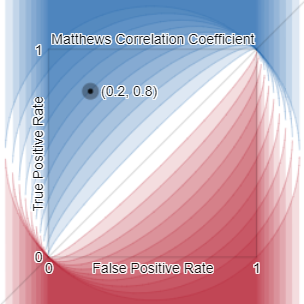}
\caption{MCC with Prev=0.5}
\end{subfigure}\\[2ex]
\begin{subfigure}[b]{\textwidth}
\centering
\includegraphics[height=4cm]{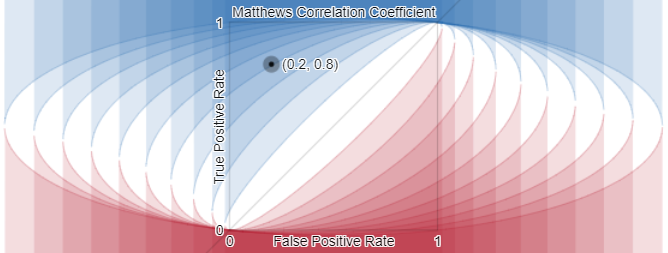}
\caption{MCC with Prev=0.9}
\end{subfigure}
\end{minipage}
\caption{Contours of the Matthews Correlation Coefficient within and beyond the ROC space describe a series of concentric ellipses whose eccentricity depends on prevalence and which intersect where $(\FPR, \TPR) = (0,0)$ and $(\FPR, \TPR) = (1,1)$}
\label{fig:MCCbeyond}
\end{figure}

Using the notation of Table~\ref{tab:confusion-matrix}, the Matthews Correlation Coefficient (Eq.~\eqref{eq:MCC}) can be rewritten in terms of $a, d, p, n$ as
\begin{align*}
    \MCC(a, b, c, d)
    &=\frac{ad-bc}{\sqrt{(a+b)(a+c)(b+d)(c+d)}}\\
    &=\frac{ad-(n-d)(p-a)}{\sqrt{(a+n-d)pn(p-a+d)}}.
\end{align*}
For given numbers of positives ($p$) and negatives ($n$), this performance metric achieves a value of $-1 \leq k \leq 1$ along the contour lines with
\begin{multline}
a(k, p, n, d) = \\
\begin{dcases}
\frac{1}{2 (k^2 p + n)}
\left( 
  +\sqrt{
    \frac
      {k^2 p (n + p)^2 (4d(n-d) + k^2 n p)}
      {n}
  } + 
  2dp(k^2 - 1) + k^2p(p - n) + 2np
\right) & k \geq 0\\
\frac{1}{2 (k^2 p + n)}
\left( 
  -\sqrt{
    \frac
      {k^2 p (n + p)^2 (4d(n-d) + k^2 n p)}
      {n}
  } + 
  2dp(k^2 - 1) + k^2p(p - n) + 2np
\right) & k < 0
\end{dcases}
\end{multline}
or, in terms of true positive rate $\alpha=a/p$ and true negative rate $\delta=d/n$
\begin{multline}
\alpha(k, p, n, \delta) = \\
\begin{dcases}
\frac{1}{2p (k^2 p + n)}
\left( 
  +\sqrt{
    \frac
      {k^2 p (n + p)^2 (4n^2\delta(n-\delta) + k^2 n p)}
      {n}
  } + 
  2n\delta p(k^2 - 1) + k^2p(p - n) + 2np
\right) & k \geq 0\\
\frac{1}{2p (k^2 p + n)}
\left( 
  -\sqrt{
    \frac
      {k^2 p (n + p)^2 (4n^2\delta(n-\delta) + k^2 n p)}
      {n}
  } + 
  2n\delta p(k^2 - 1) + k^2p(p - n) + 2np
\right) & k < 0
\end{dcases}
\end{multline}
in which case, all contours where $k < 0$ intersect at $(\alpha,\delta)=(0,1)$ and all contours where $k > 0$ intersect at $(\alpha,\delta)=(1,0)$, as is the case for Markedness. These intersections become apparent when we visualise the contours of the Matthews Correlation Coefficient within and beyond the ROC space (Figure~\ref{fig:MCCbeyond}) which reveals that the contours describe a series of concentric ellipses whose eccentricity depends on prevalence.

\clearpage
\subsection{\texorpdfstring{$\fOne$}{F1} Score}
\label{sec:F1}


\begin{figure}
\begin{subfigure}[b]{0.33\textwidth}
    \centering
    \includegraphics[width=\textwidth]{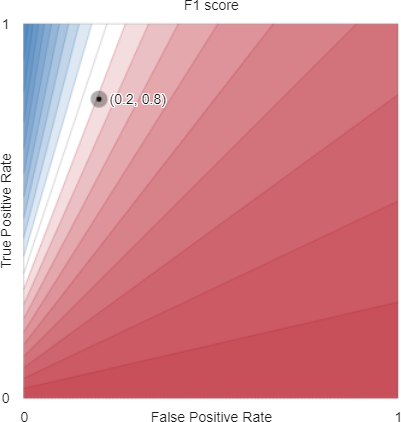}
    \caption{$\fOne$ score with Prev = 0.1}
    \label{fig:F1.p10}
\end{subfigure}
\begin{subfigure}[b]{0.33\textwidth}
    \centering
    \includegraphics[width=\textwidth]{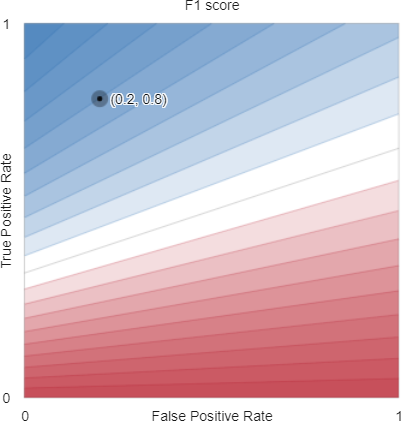}
    \caption{$\fOne$ score with Prev = 0.5}
    \label{fig:F1.p50}
\end{subfigure}
\begin{subfigure}[b]{0.33\textwidth}
    \centering
    \includegraphics[width=\textwidth]{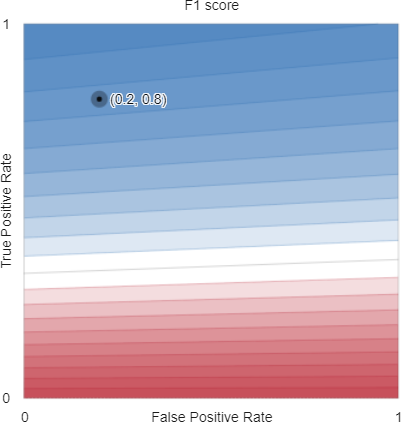}
    \caption{$\fOne$ score with Prev = 0.9}
    \label{fig:F1.p90}
\end{subfigure}
\caption{Contours of the $\fOne$ score in the ROC space.}
\label{fig:F1.contours}
\end{figure}

\begin{figure}[t]
\begin{subfigure}[b]{\textwidth}
\centering
\includegraphics[width=\textwidth]{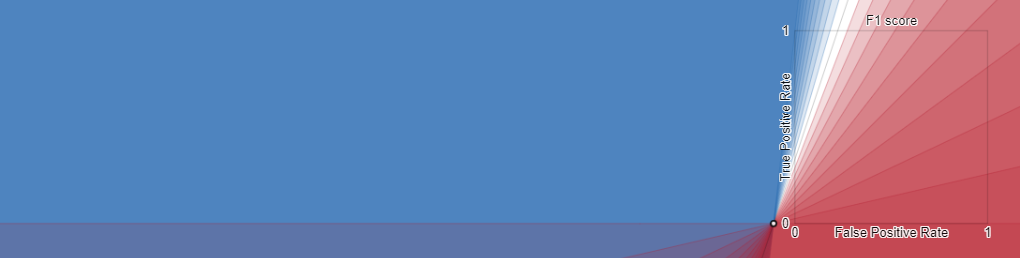}
\caption{$\fOne$ with Prev=0.1}
\end{subfigure}\\[2ex]
\begin{subfigure}[b]{\textwidth}
\centering
\includegraphics[width=\textwidth]{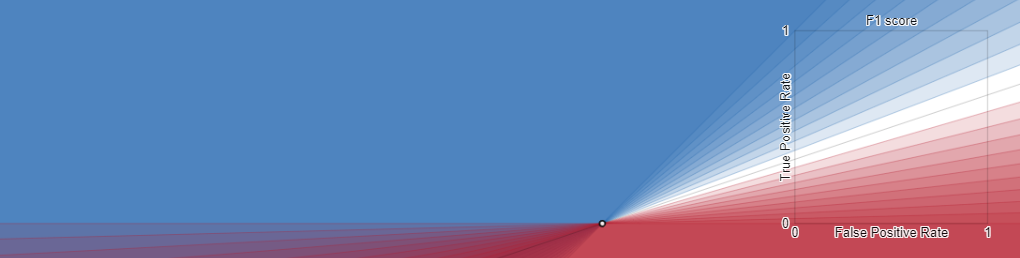}
\caption{$\fOne$ with Prev=0.5}
\end{subfigure}\\[2ex]
\begin{subfigure}[b]{\textwidth}
\centering
\includegraphics[width=\textwidth]{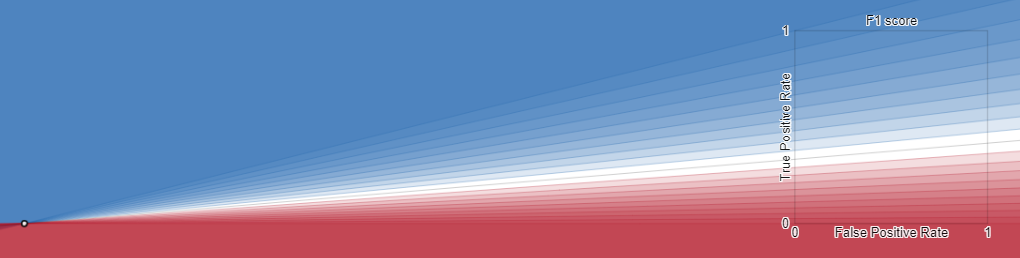}
\caption{$\fOne$ with Prev=0.8}
\end{subfigure}
\caption{Contours of the $\fOne$ score in and beyond the ROC space are straight lines that intersect at $(\alpha,\delta)=(0,(p+n)/n)$ or, equivalently $(\FPR, \TPR)=(1-(p+n)/n, 0)$.}
\label{fig:F1beyond}
\end{figure}

Using the notation of Table~\ref{tab:confusion-matrix}, the $\fOne$ score (Eq.~\eqref{eq:F1}) can be rewritten in terms of $a, d, p, n$ as
\begin{align*}
    \fOne(a, b, c, d)
    &=\frac{2a}{2a + b + c}\\
    &=\frac{2a}{2 a + n - d + p - a}.
\end{align*}
For given numbers of positives ($p$) and negatives ($n$), this performance metric achieves a value of $0 \leq k \leq 1$ along the contour lines with
\begin{align}
a(k, p, n, d) = \frac{k (d - n - p)}{k - 2}
\end{align}
or, in terms of true positive rate $\alpha=a/p$ and true negative rate $\delta=d/n$
\begin{align}
\alpha(k, p, n, \delta) = 
\frac{k (n\delta - n - p)}{p(k - 2)}
\end{align}
in which case, all contours intersect at $(\alpha,\delta)=(0,(p+n)/n)$. These intersections become apparent when we visualise the contours of the $\fOne$ score within and beyond the ROC space (Figure~\ref{fig:F1beyond}) which reveals how the slopes of the linear contours depend on prevalence, similar to those of the Threat Score.

\clearpage
\subsection{Threat Score (Jaccard Index, Critical Success Index)}

\begin{figure}
\begin{subfigure}[b]{0.33\textwidth}
    \centering
    \includegraphics[width=\textwidth]{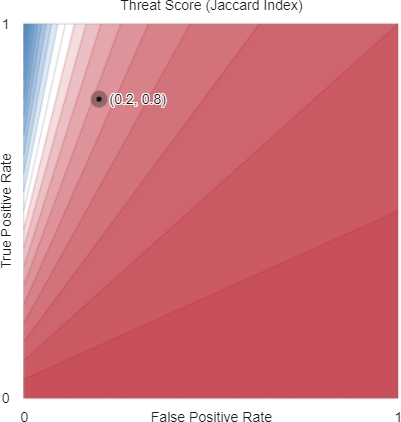}
    \caption{Threat Score with Prev = 0.1}
    \label{fig:TS.10}
\end{subfigure}
\begin{subfigure}[b]{0.33\textwidth}
    \centering
    \includegraphics[width=\textwidth]{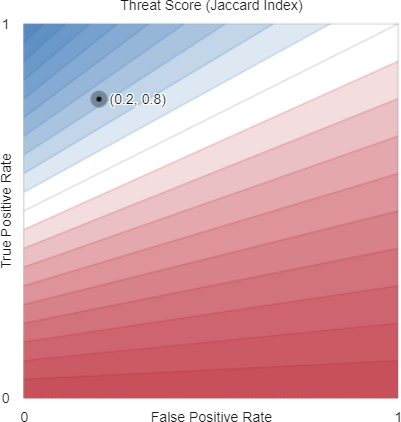}
    \caption{Threat Score with Prev = 0.5}
    \label{fig:TS.50}
\end{subfigure}
\begin{subfigure}[b]{0.33\textwidth}
    \centering
    \includegraphics[width=\textwidth]{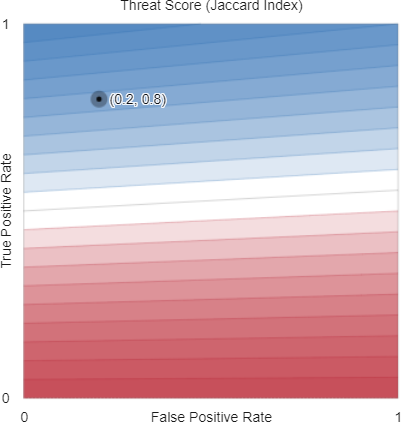}
    \caption{Threat Score with Prev = 0.9}
    \label{fig:TS.90}
\end{subfigure}
\caption{Contours of the Threat Score (Jaccard Index, Critical Success Index) in the ROC space.}
\label{fig:TS.contours}
\end{figure}

\begin{figure}[t]
\begin{subfigure}[b]{\textwidth}
\centering
\includegraphics[width=\textwidth]{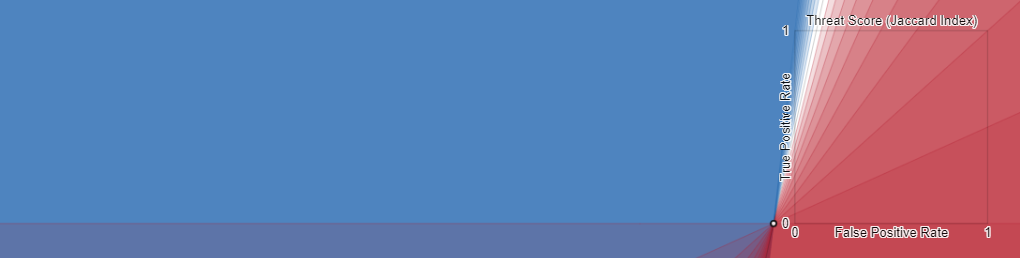}
\caption{Threat Score with Prev=0.1}
\end{subfigure}\\[2ex]
\begin{subfigure}[b]{\textwidth}
\centering
\includegraphics[width=\textwidth]{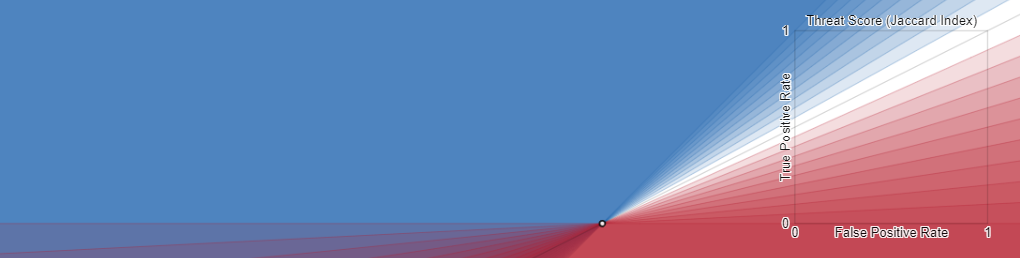}
\caption{Threat Score with Prev=0.5}
\end{subfigure}\\[2ex]
\begin{subfigure}[b]{\textwidth}
\centering
\includegraphics[width=\textwidth]{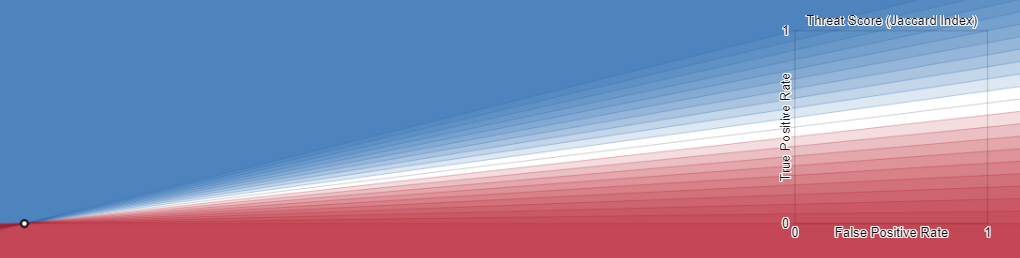}
\caption{Threat Score with Prev=0.8}
\end{subfigure}
\caption{Contours of the Threat Score score in and beyond the ROC space are straight lines that intersect at $(\alpha,\delta)=(0,(p+n)/n)$ or, equivalently $(\FPR, \TPR)=(1-(p+n)/n, 0)$.}
\label{fig:TSbeyond}
\end{figure}

Using the notation of Table~\ref{tab:confusion-matrix}, the Threat Score (Eq.~\eqref{eq:TS}) can be rewritten in terms of $a, d, p, n$ as
\begin{align*}
    \TS(a, b, c, d)
    &=\frac{a}{a + b + c}\\
    &=\frac{a}{p + n - d}.
\end{align*}
For given numbers of positives ($p$) and negatives ($n$), this performance metric achieves a value of $0 \leq k \leq 1$ along the contour lines with
\begin{align}
a(k, p, n, d) = k(p + n - d)
\end{align}
or, in terms of true positive rate $\alpha=a/p$ and true negative rate $\delta=d/n$
\begin{align}
\alpha(k, p, n, \delta) = 
k(p + n(1 - \delta))/p
\end{align}
in which case, all contours intersect at $(\alpha,\delta)=(0,(p+n)/n)$.
These intersections become apparent when we visualise the contours of the $\fOne$ score within and beyond the ROC space (Figure~\ref{fig:TSbeyond}) which reveals how the slopes of the linear contours depend on prevalence, similar to those of the $\fOne$ Score.

\clearpage
\subsection{Accuracy}

\begin{figure}
\begin{subfigure}[b]{0.33\textwidth}
    \centering
    \includegraphics[width=\textwidth]{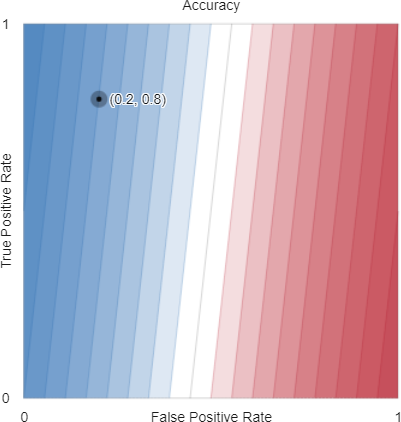}
    \caption{Accuracy with Prev = 0.1}
    \label{fig:Acc.10}
\end{subfigure}
\begin{subfigure}[b]{0.33\textwidth}
    \centering
    \includegraphics[width=\textwidth]{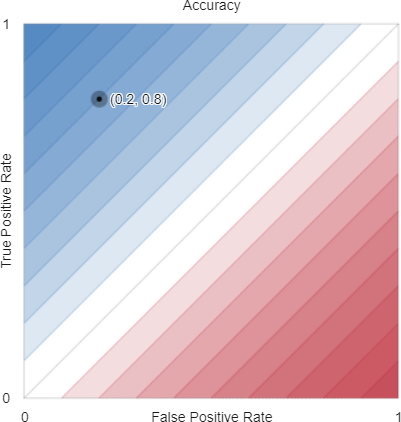}
    \caption{Accuracy with Prev = 0.5}
    \label{fig:Acc.50}
\end{subfigure}
\begin{subfigure}[b]{0.33\textwidth}
    \centering
    \includegraphics[width=\textwidth]{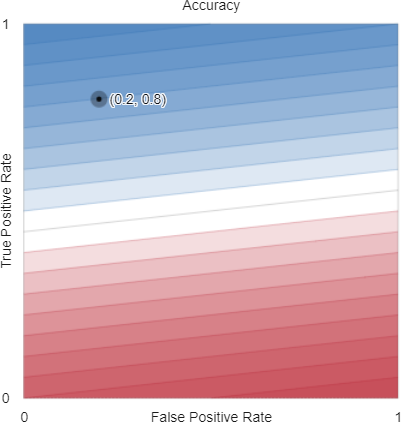}
    \caption{Accuracy with Prev = 0.9}
    \label{fig:Acc.90}
\end{subfigure}
\caption{Contours of accuracy in the ROC space.}
\label{fig:Acc.contours}
\end{figure}


Using the notation of Table~\ref{tab:confusion-matrix}, the Accuracy (Eq.~\eqref{eq:Acc}) can be rewritten in terms of $a, d, p, n$ as
\begin{align*}
    \TS(a, b, c, d)
    &=\frac{a + d}{a + b + c+d}\\
    &=\frac{a+d}{p + n}.
\end{align*}
For given numbers of positives ($p$) and negatives ($n$), this performance metric achieves a value of $0 \leq k < 1$ along the contour lines with
\begin{align}
a(k, p, n, d) = k(p + n) - d
\end{align}
or, in terms of true positive rate $\alpha=a/p$ and true negative rate $\delta=d/n$
\begin{align}
\alpha(k, p, n, \delta) = 
\frac{k(p + n) - n\delta}{p}.
\end{align}
These contour lines are parallel and planar.
\clearpage

\subsection{Decision Costs (Benefits)}
\label{sec:decision-costs}

\begin{figure}
\begin{subfigure}[b]{0.33\textwidth}
    \centering
    \includegraphics[width=\textwidth]{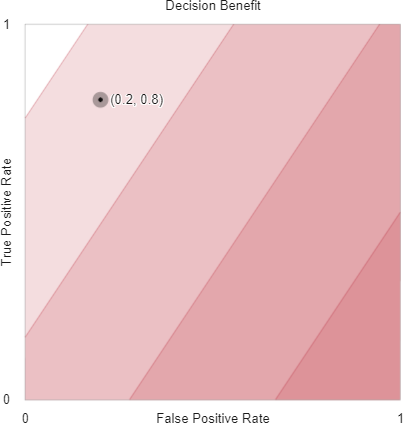}
    \caption{Decision Benefits with Prev = 0.1}
    \label{fig:DB.7.3.1.4.10}
\end{subfigure}
\begin{subfigure}[b]{0.33\textwidth}
    \centering
    \includegraphics[width=\textwidth]{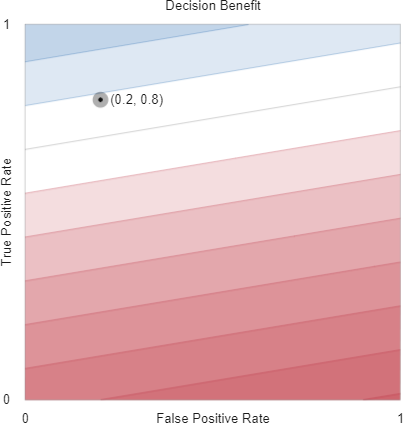}
    \caption{Decision Benefits  with Prev = 0.5}
    \label{fig:DB.7.3.1.4.50}
\end{subfigure}
\begin{subfigure}[b]{0.33\textwidth}
    \centering
    \includegraphics[width=\textwidth]{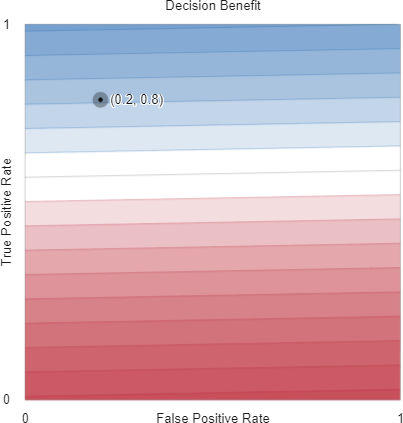}
    \caption{Decision Benefits  with Prev = 0.9}
    \label{fig:DB.7.3.1.4.90}
\end{subfigure}
\caption{Contours of Decision Benefits in the ROC space using $\benefits=\SmallMatrix{7&3\\1&4}$.}
\label{fig:DB.contours1}
\vspace{\textfloatsep}
\begin{subfigure}[b]{0.33\textwidth}
    \centering
    \includegraphics[width=\textwidth]{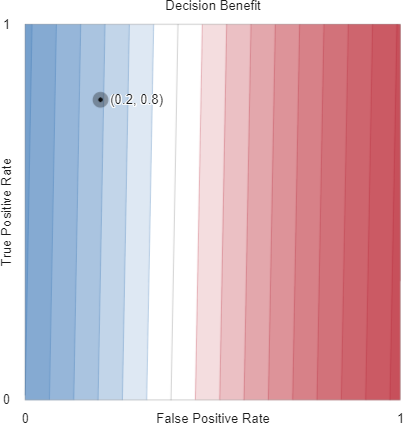}
    \caption{Decision Benefits with Prev = 0.1}
    \label{fig:DB.10}
\end{subfigure}
\begin{subfigure}[b]{0.33\textwidth}
    \centering
    \includegraphics[width=\textwidth]{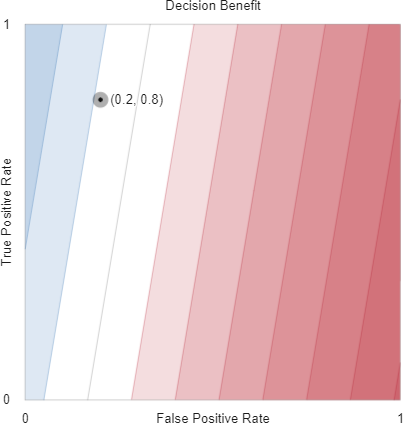}
    \caption{Decision Benefits  with Prev = 0.5}
    \label{fig:DB.50}
\end{subfigure}
\begin{subfigure}[b]{0.33\textwidth}
    \centering
    \includegraphics[width=\textwidth]{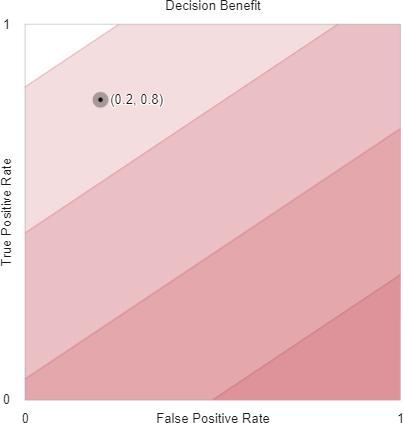}
    \caption{Decision Benefits  with Prev = 0.9}
    \label{fig:DB.90}
\end{subfigure}
\caption{Contours of Decision Benefits in the ROC space using $\benefits=\SmallMatrix{4&1\\1&7}$.}
\label{fig:DB.contours2}
\end{figure}

While it is common to refer to the \textit{costs} of correct and incorrect classifications, we work here equivalently in terms of \textit{benefits} in keeping with other performance metrics in this paper where ``bigger is better'' and to ensure the colour scales used in our graphics can be interpreted consistently (``blue is better'').

We define the matrix of benefits associated with the confusion matrix of Table~\ref{tab:confusion-matrix} as
\begin{align*}
\benefits &=
    \begin{bmatrix}
    \beta_a & \beta_b\\
    \beta_c & \beta_d
    \end{bmatrix}.
\end{align*}
Literature that concentrates on decision costs for a \textit{fixed prevalence} treats this matrix as having only two degrees of freedom, e.g., the ratio (or difference) of costs for true positive and false negative classifications, and the ratio (or difference) of costs for true negative and false positive classifications. This paper considers confusion matrices with fixed totals, but \textit{varying prevalence}, so our parameterisation of the benefits matrix affords three degrees of freedom.

Using the notation of Table~\ref{tab:confusion-matrix}, the Decision Benefits (Eq.~\eqref{eq:DB}) can be rewritten in terms of $a, d, p, n$ and benefits $\benefits$ as
\begin{align*}
    \DB(a, b, c, d, \benefits)
    &=a\cdot\beta_a + b\cdot\beta_b + c\cdot\beta_c + d\cdot\beta_d\\
    &=a\cdot\beta_a + (n-d)\beta_b + (p-a)\beta_c + d\cdot\beta_d\\
    &=a(\beta_a - \beta_c) + d(\beta_d - \beta_b) + p\cdot\beta_c + n\cdot\beta_d
\end{align*}
For given numbers of positives ($p$) and negatives ($n$), this performance metric achieves a value of $k$ along the contour lines with
\begin{align}
a(k, p, n, d, \benefits) = 
-d\frac{\beta_d - \beta_b}{\beta_a - \beta_c} + \frac{k-p\cdot\beta_c - n\cdot\beta_d}{\beta_a - \beta_c}
\end{align}
or, in terms of true positive rate $\alpha=a/p$ and true negative rate $\delta=d/n$
\begin{align}
\alpha(k, p, n, \delta,\benefits) = 
-\delta \frac{n(\beta_d - \beta_b)}{p(\beta_a - \beta_c)} + \frac{k-p\cdot\beta_c - n\cdot\beta_d}{p(\beta_a - \beta_c)}.
\end{align}
These contour lines are parallel and planar.

To help with plotting the contours for all confusion matrices of size $\Tot$, we use a shifted and scaled version of the benefits matrix, $\benefits^\ast$, whose minimum element is 0 and whose maximum element is $\tfrac{1}{\Tot}$
\begin{align*}
    \benefits^\ast = \frac{1}{\Tot}\cdot\frac{\benefits - \min(\benefits)}{\max(\benefits)}.
\end{align*}
This ensures feasible contours range between $[0,1]$. In interactive plotting, we enforce the constraints
\begin{align*}
    \beta_a > \beta_c &\geq 0\\
    \beta_d > \beta_b &\geq 0
\end{align*}
to ensure that the contours have finite, positive slope.

When
\begin{align*}
    \benefits &= 
    \begin{bmatrix}
    \beta & 0\\
    0 & \beta
    \end{bmatrix}
\end{align*}
for some positive $\beta$, the Decision Cost contours are the same as those of Accuracy.

\clearpage
\subsection{Positive Predictive Value (Precision)}

\begin{figure}
\begin{subfigure}[b]{0.33\textwidth}
    \centering
    \includegraphics[width=\textwidth]{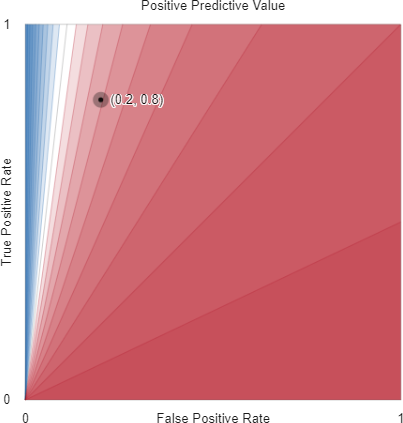}
    \caption{Positive Predictive Value with Prev = 0.1}
    \label{fig:PPV.10}
\end{subfigure}
\begin{subfigure}[b]{0.33\textwidth}
    \centering
    \includegraphics[width=\textwidth]{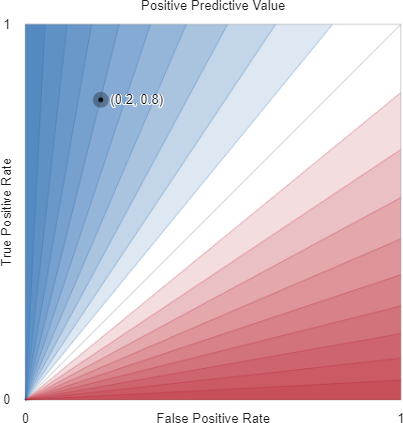}
    \caption{Positive Predictive Value with Prev = 0.5}
    \label{fig:PPV.50}
\end{subfigure}
\begin{subfigure}[b]{0.33\textwidth}
    \centering
    \includegraphics[width=\textwidth]{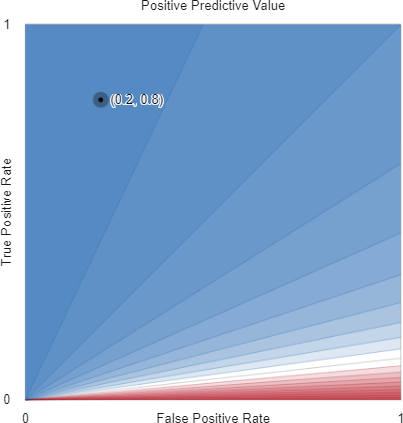}
    \caption{Positive Predictive Value with Prev = 0.9}
    \label{fig:PPV.90}
\end{subfigure}
\caption{Contours of Positive Predictive Value in the ROC space.}
\label{fig:PPV.contours}
\end{figure}


Using the notation of Table~\ref{tab:confusion-matrix}, the Positive Predictive Value (Eq.~\eqref{eq:PPV}) can be rewritten in terms of $a, d, p, n$ as
\begin{align*}
    \PPV(a, b, c, d)
    &=\frac{a}{a + b}\\
    &=\frac{a}{a + n - d}.
\end{align*}
For given numbers of positives ($p$) and negatives ($n$), this performance metric achieves a value of $0 \leq k \leq 1$ along the contour lines with
\begin{align}
a(k, p, n, d) = \frac{k(d-n)}{k-1}
\end{align}
or, in terms of true positive rate $\alpha=a/p$ and true negative rate $\delta=d/n$
\begin{align}
\alpha(k, p, n, \delta) = 
\frac{kn(\delta-1)}{p(k-1)}
\end{align}
in which case, all contours intersect at $(\alpha,\delta)=(0,1)$.

\clearpage

\subsection{Negative Predictive Value}

\begin{figure}
\begin{subfigure}[b]{0.33\textwidth}
    \centering
    \includegraphics[width=\textwidth]{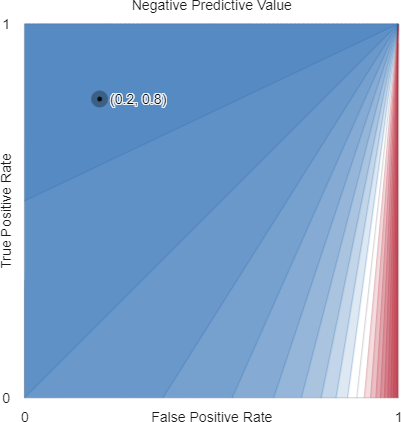}
    \caption{Negative Predictive Value with Prev = 0.1}
    \label{fig:NPV.10}
\end{subfigure}
\begin{subfigure}[b]{0.33\textwidth}
    \centering
    \includegraphics[width=\textwidth]{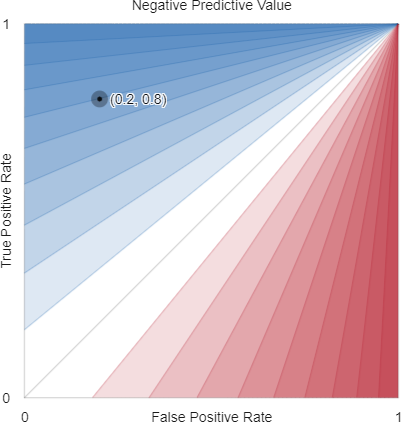}
    \caption{Negative Predictive Value with Prev = 0.5}
    \label{fig:NPV.50}
\end{subfigure}
\begin{subfigure}[b]{0.33\textwidth}
    \centering
    \includegraphics[width=\textwidth]{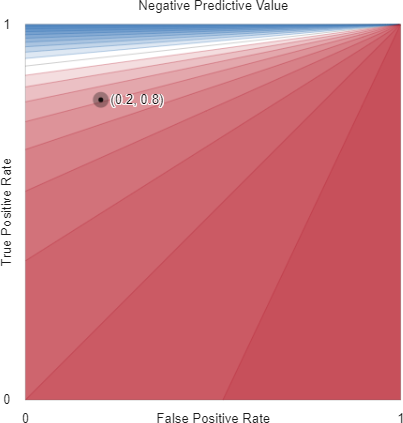}
    \caption{Negative Predictive Value with Prev = 0.9}
    \label{fig:NPV.90}
\end{subfigure}
\caption{Contours of Negative Predictive Value in the ROC space.}
\label{fig:NPV.contours}
\end{figure}


Using the notation of Table~\ref{tab:confusion-matrix}, the Negative Predictive Value (Eq.~\eqref{eq:PPV}) can be rewritten in terms of $a, d, p, n$ as
\begin{align*}
    \NPV(a, b, c, d)
    &=\frac{d}{c+ d}\\
    &=\frac{d}{p-a + d}.
\end{align*}
For given numbers of positives ($p$) and negatives ($n$), this performance metric achieves a value of $0 \leq k \leq 1$ along the contour lines with
\begin{align}
a(k, p, n, d) = p +  \frac{d(k-1)}{k}
\end{align}
or, in terms of true positive rate $\alpha=a/p$ and true negative rate $\delta=d/n$
\begin{align}
\alpha(k, p, n, \delta) = 
1 +  \frac{n\delta(k-1)}{pk}
\end{align}
in which case, all contours intersect at $(\alpha,\delta)=(1,0)$.

\clearpage

\subsection{Markedness}

\begin{figure}
\begin{subfigure}[b]{0.33\textwidth}
    \centering
    \includegraphics[width=\textwidth]{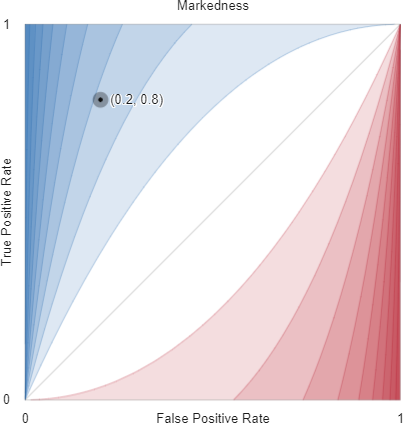}
    \caption{Markedness with Prev = 0.1}
    \label{fig:MK.10}
\end{subfigure}
\begin{subfigure}[b]{0.33\textwidth}
    \centering
    \includegraphics[width=\textwidth]{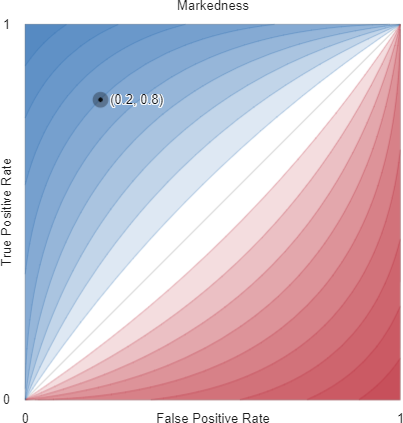}
    \caption{Markedness with Prev = 0.5}
    \label{fig:MK.50}
\end{subfigure}
\begin{subfigure}[b]{0.33\textwidth}
    \centering
    \includegraphics[width=\textwidth]{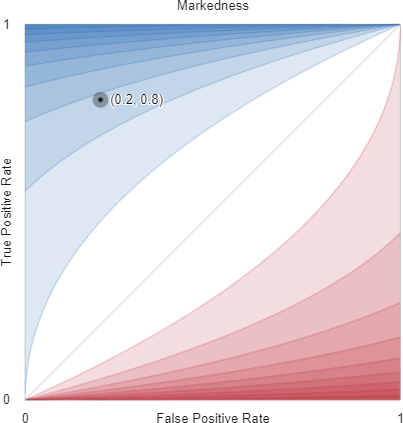}
    \caption{Markedness with Prev = 0.9}
    \label{fig:MK.90}
\end{subfigure}
\caption{Contours of Markedness in the ROC space.}
\label{fig:MK.contours}
\end{figure}


Using the notation of Table~\ref{tab:confusion-matrix},  Markedness (Eq.~\eqref{eq:MK}) can be rewritten in terms of $a, d, p, n$ as
\begin{align*}
    \MK(a, b, c, d)
    &=\frac{a}{a + b} + \frac{d}{c+ d} - 1\\
    &=\frac{a}{a + n-d} + \frac{d}{p-a + d} - 1.
\end{align*}
For given numbers of positives ($p$) and negatives ($n$), this performance metric achieves a value of $0 \leq k \leq 1$ along the contour lines with
\begin{align}
a(k, p, n, d) = 
\begin{dcases}
\frac{\sqrt{(kn+kp+n)^2-4dk(n+p)} + 2dk - kn + kp -n}{2k}
 & k \geq 0\\
-\frac{\sqrt{(kn+kp+n)^2-4dk(n+p)} - 2dk + kn - kp +n}{2k}
 & k < 0
\end{dcases}
\end{align}
or, in terms of true positive rate $\alpha=a/p$ and true negative rate $\delta=d/n$
\begin{align}
\alpha(k, p, n, \delta) = 
\begin{dcases}
\frac{\sqrt{(kn+kp+n)^2-4n\delta k(n+p)} + 2n\delta k - kn + kp -n}{2pk}
 & k \geq 0\\
-\frac{\sqrt{(kn+kp+n)^2-4n\delta k(n+p)} - 2n\delta k + kn - kp +n}{2pk}
 & k < 0
\end{dcases}
\end{align}
in which case, all contours where $k < 0$ intersect at $(\alpha,\delta)=(0,1)$ and all contours where $k > 0$ intersect at $(\alpha,\delta)=(1,0)$, as is the case for the Matthews Correlation Coefficient.

\begin{figure}[t]
\begin{subfigure}[b]{4.5cm}
\centering
\includegraphics[width=4cm]{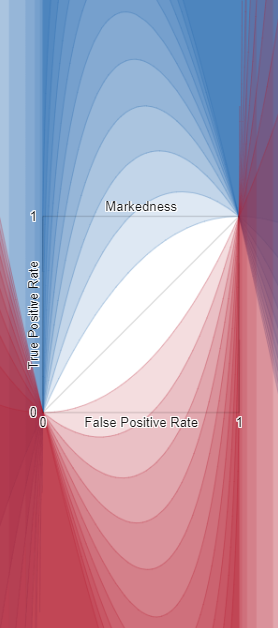}
\caption{Markedness with Prev=0.1}
\end{subfigure}
\begin{minipage}[b]{\dimexpr\textwidth-5cm}
\begin{subfigure}[b]{\textwidth}
\centering
\includegraphics[width=5cm]{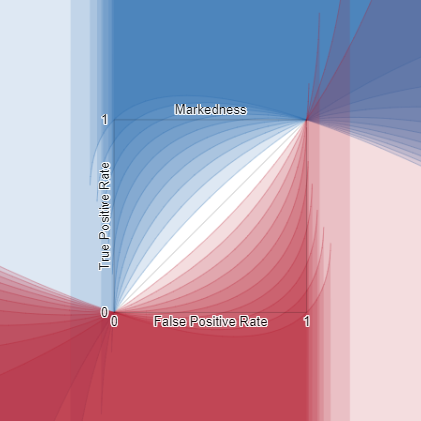}
\caption{Markedness with Prev=0.5}
\end{subfigure}\\[2ex]
\begin{subfigure}[b]{\textwidth}
\centering
\includegraphics[height=4cm,trim={3.7cm 0 3.7cm 0},clip]{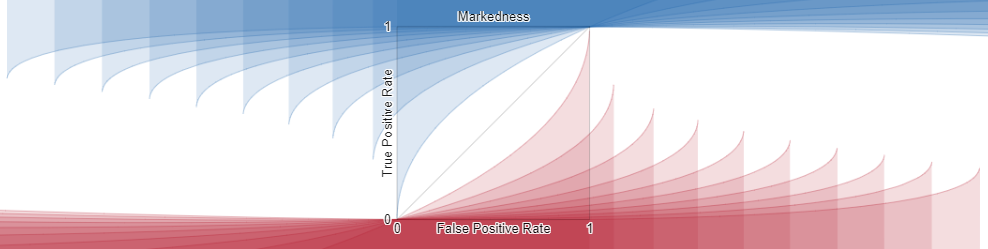}
\caption{Markedness with Prev=0.9}
\end{subfigure}
\end{minipage}
\caption{Contours of Markedness within and beyond the ROC space describe a series of polynomial curves whose shape depends on prevalence and which intersect where $(\FPR, \TPR) = (0,0)$ and $(\FPR, \TPR) = (1,1)$}
\label{fig:MKbeyond}
\end{figure}

\clearpage
\subsection{Cohen's kappa}

\begin{figure}
\begin{subfigure}[b]{0.33\textwidth}
    \centering
    \includegraphics[width=\textwidth]{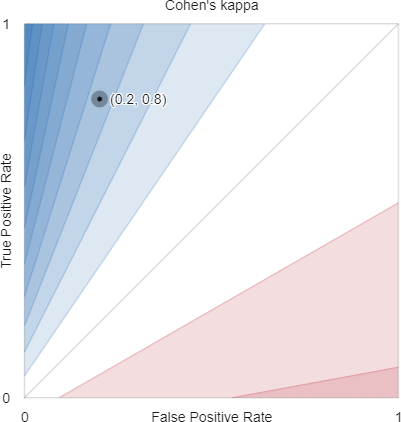}
    \caption{Cohen's kappa with Prev = 0.1}
    \label{fig:CK.10}
\end{subfigure}
\begin{subfigure}[b]{0.33\textwidth}
    \centering
    \includegraphics[width=\textwidth]{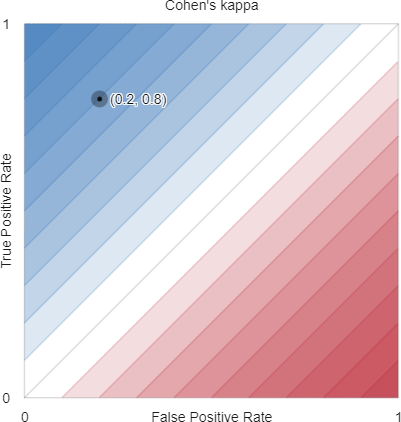}
    \caption{Cohen's kappa with Prev = 0.5}
    \label{fig:CK.50}
\end{subfigure}
\begin{subfigure}[b]{0.33\textwidth}
    \centering
    \includegraphics[width=\textwidth]{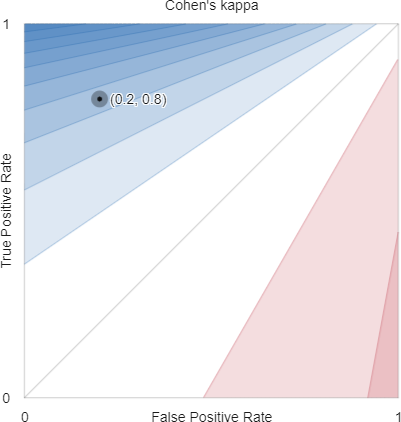}
    \caption{Cohen's kappa with Prev = 0.9}
    \label{fig:CK.90}
\end{subfigure}
\caption{Contours of Cohen's kappa in the ROC space.}
\label{fig:CK.contours}
\end{figure}


Using the notation of Table~\ref{tab:confusion-matrix}, Cohen's kappa (Eq.~\eqref{eq:CK}) can be rewritten in terms of $a, d, p, n$ as
\begin{align*}
    \CK(a, b, c, d)
    &=\frac{2(ad - bc)}{(a + b)(b + d) + (a + c)(c + d)}\\
    &=\frac{2(ad - (n-d)(p-a))}{(a + n - d)n + p(p - a + d)}.
\end{align*}
For given numbers of positives ($p$) and negatives ($n$), this performance metric achieves a value of $0 \leq k \leq 1$ along the contour lines with
\begin{align}
a(k, p, n, d) =  \frac{d k (n - p) + 2 d p - k (n^2 + p^2) - 2 n p}{(k - 2) n - k p}
\end{align}
or, in terms of true positive rate $\alpha=a/p$ and true negative rate $\delta=d/n$
\begin{align}
\alpha(k, p, n, \delta) &= 
\frac{n \delta k (n - p) + 2 n \delta p - k (n^2 + p^2) - 2 n p}{p((k - 2) n - k p)}\\
&=
\frac{n \delta k (n - p) - 2 n p (1 - \delta )  - k (n^2 + p^2)}{p(k(n-p) - 2n)}
\end{align}
in which case, all contours intersect at
\[
(\alpha,\delta)=\left(\frac{n}{n-p},1-\frac{n}{n-p}\right)
\]
when $p \neq n$.


\clearpage
\subsection{Fowlkes-Mallows index}

\begin{figure}
\begin{subfigure}[b]{0.33\textwidth}
    \centering
    \includegraphics[width=\textwidth]{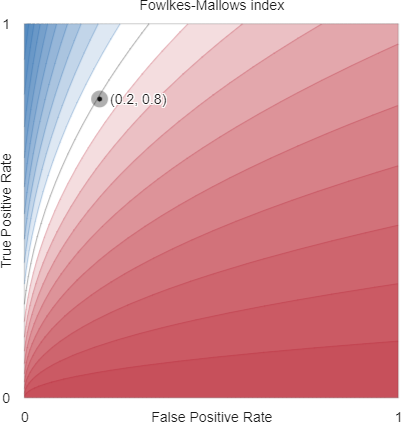}
    \caption{Fowlkes-Mallows index with Prev = 0.1}
    \label{fig:FM.10}
\end{subfigure}
\begin{subfigure}[b]{0.33\textwidth}
    \centering
    \includegraphics[width=\textwidth]{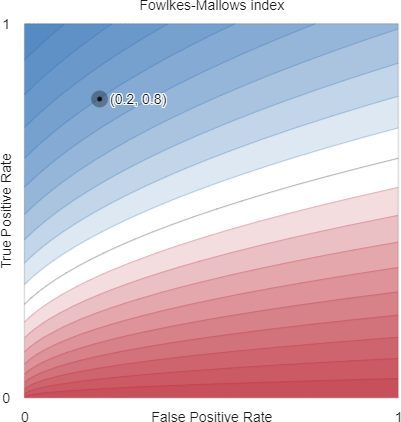}
    \caption{Fowlkes-Mallows index with Prev = 0.5}
    \label{fig:FM.50}
\end{subfigure}
\begin{subfigure}[b]{0.33\textwidth}
    \centering
    \includegraphics[width=\textwidth]{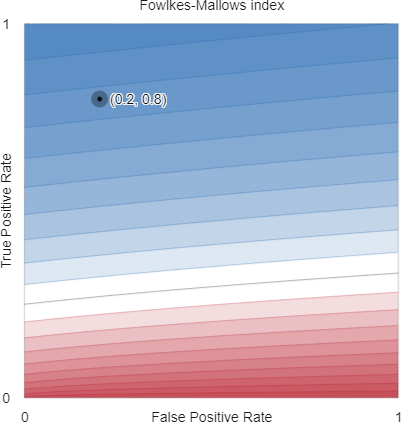}
    \caption{Fowlkes-Mallows index with Prev = 0.9}
    \label{fig:FM.90}
\end{subfigure}
\caption{Contours of Fowlkes-Mallows index in the ROC space.}
\label{fig:FM.contours}
\end{figure}


Using the notation of Table~\ref{tab:confusion-matrix}, the Fowlkes-Mallows Index  (Eq.~\eqref{eq:FM}) can be rewritten in terms of $a, d, p, n$ as
\begin{align*}
    \FM(a, b, c, d)
    &=\sqrt{\frac{a}{a+b}\cdot\frac{a}{a+c}}\\
    &=\sqrt{\frac{a}{a+n-d}\cdot\frac{a}{p}}
\end{align*}
For given numbers of positives ($p$) and negatives ($n$), this performance metric achieves a value of $0 \leq k \leq 1$ along the contour lines with
\begin{align}
a(k, p, n, d) = \frac{1}{2} \left(\sqrt{k^2 p (-4 d + k^2 p + 4 n)} + k^2 p\right)
\end{align}
or, in terms of true positive rate $\alpha=a/p$ and true negative rate $\delta=d/n$
\begin{align}
\alpha(k, p, n, \delta) = 
\frac{1}{2p}\left(\sqrt{k^{2}p(-4nd+k^{2}p+4n)}+k^{2}p\right).
\end{align}

\clearpage 

\section{Performance metric contours that are independent of prevalence}
\label{sec:prevalence-independent-contours}

\begin{SCfigure}[50][t]
  \caption{Contours of True Positive Rate in the ROC space.}
  \includegraphics[width=0.3\textwidth]{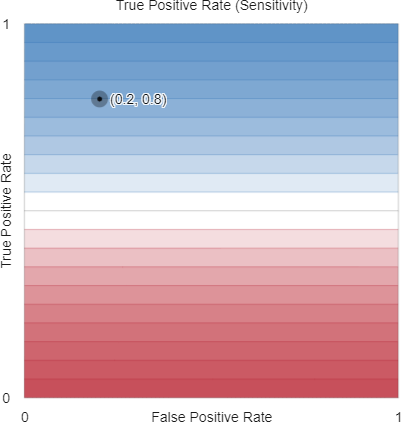}
  \label{fig:TPR}
\end{SCfigure}

\subsection{True Positive Rate (Sensitivity)}

Using the notation of Table~\ref{tab:confusion-matrix}, the True Positive Rate (Eq.~\eqref{eq:TPR}) can be rewritten in terms of $a, d, p, n$ as
\begin{align*}
    \TPR(a, b, c, d)
    &=\frac{a}{a + c}\\
    &=\frac{a}{p}.
\end{align*}
For given numbers of positives ($p$), this performance metric achieves a value of $0 \leq k \leq 1$ along the contour lines with
\begin{align}
a(k, p) = pk
\end{align}
or, in terms of true positive rate $\alpha=a/p$ and true negative rate $\delta=d/n$
\begin{align}
\alpha(k) =  k
\end{align}

\clearpage 

\begin{SCfigure}[50][t]
  \caption{Contours of True Negative Rate in the ROC space.}
  \includegraphics[width=0.3\textwidth]{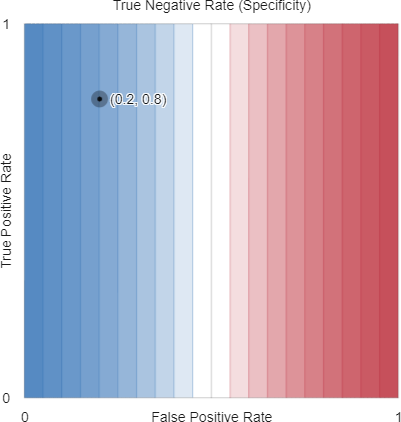}
  \label{fig:TNR}
\end{SCfigure}

\subsection{True Negative Rate (Specificity)}

Using the notation of Table~\ref{tab:confusion-matrix}, the True Positive Rate (Eq.~\eqref{eq:TNR}) can be rewritten in terms of $a, d, p, n$ as
\begin{align*}
    \TPR(a, b, c, d)
    &=\frac{d}{b + d}\\
    &=\frac{d}{n}.
\end{align*}
For given numbers of negatives ($n$), this performance metric achieves a value of $0 \leq k \leq 1$ along the contour lines with
\begin{align}
d(k, n) = nk
\end{align}
or, in terms of true negative rate $\delta=d/n$
\begin{align}
\delta(k) =  k
\end{align}

\clearpage 

\begin{SCfigure}[50]
  \caption{Contours of the geometric mean of true positive rate (sensitivity) and true negative rate (specificity) in the ROC space.}
  \includegraphics[width=0.3\textwidth]{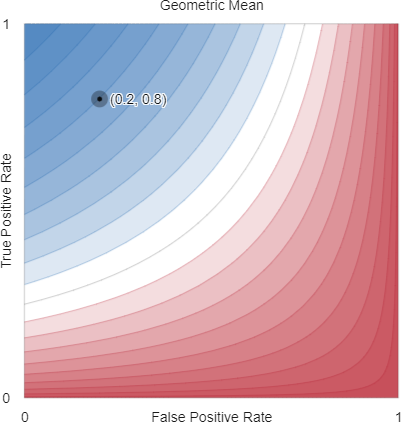}
  \label{fig:GM}
\end{SCfigure}

\subsection{Geometric Mean}


Using the notation of Table~\ref{tab:confusion-matrix}, the True Positive Rate (Eq.~\eqref{eq:TNR}) can be rewritten in terms of $a, d, p, n$ as
\begin{align*}
    \GM(a, b, c, d)
    &=\sqrt{\frac{a}{a+c}\cdot\frac{d}{b+d}}\\
    &=\sqrt{\frac{a}{p}\cdot\frac{d}{n}}.
\end{align*}
For given numbers of positives ($p$) and negatives ($n$), this performance metric achieves a value of $0 \leq k \leq 1$ along the contour lines with
\begin{align}
a(k, p, n, d) = \frac{k^2np}{d}
\end{align}
or, in terms of true positive rate $\alpha=a/p$ and true negative rate $\delta=d/n$
\begin{align}
\alpha(k, \delta) = 
\frac{k^2}{\delta}
\end{align}

\clearpage 

\begin{SCfigure}[50]
  \caption{Contours of the scaled logarithm of the positive likelihood ratio in the ROC space.}
  \includegraphics[width=0.3\textwidth]{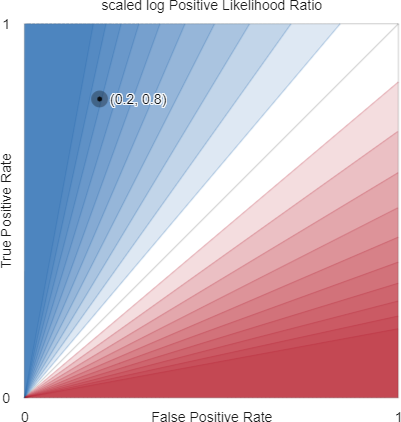}
  \label{fig:slogPLR}
\end{SCfigure}

\subsection{Positive Likelihood Ratio \texorpdfstring{($\LRP$)}{} and scaled log Positive Likelihood Ratio}
\label{sec:slLRP}


Using the notation of Table~\ref{tab:confusion-matrix}, the logarithm of the Likelihood Ratio of a positive outcomes  (Eq.~\eqref{eq:LRP}) can be rewritten in terms of $a, d, p, n$ as
\begin{align*}
    \log\LRP(a, b, c, d)
    &=\log\left(\frac{a}{p}\frac{n}{b}\right)\\
    &=\log\left(\frac{a}{p}\frac{n}{n-d}\right).
\end{align*}
For given numbers of positives ($p$) and negatives ($n$), this performance metric achieves a value of $ k $ along the contour lines with
\begin{align}
a(k, p, n, d) = \frac{e^k p (n - d)}{n} 
\label{eq:logLRP-contour}
\end{align}
or, in terms of true positive rate $\alpha=a/p$ and true negative rate $\delta=d/n$
\begin{align}
\alpha(k, \delta) = 
e^k p (1 - \delta)
\end{align}

$\log\LRP$ has range $(-\infty,\infty)$. To visualise the finite values of this function, it is convenient to work with a scaled version of this function whose contours lie between $[-1,1]$.
The largest finite value of $\LRP$ is $n(p-1)/p$ so we can produce a scaled version of Eq.~\ref{eq:logLRP-contour} by dividing $\log\LRP(a, b, c, d)$ by
\begin{align}
M = \log(n(p-1)/p)
\end{align}
to give
\begin{align}
     \text{scaled}\log\LRP(a, b, c, d)&=\frac{1}{\log(n(p-1)/p)}\log\left(\frac{a}{p}\frac{n}{b}\right)
\end{align}which yields the contour equations
\begin{align}
a(k, p, n, d) = \frac{e^{Mk} p (n - d)}{n}
\end{align}
and
\begin{align}
\alpha(k, \delta) = 
e^{Mk} p (1 - \delta)
\end{align}

\clearpage 

\begin{SCfigure}[50]
  \caption{Contours of the scaled logarithm of the negative likelihood ratio in the ROC space.}
  \includegraphics[width=0.3\textwidth]{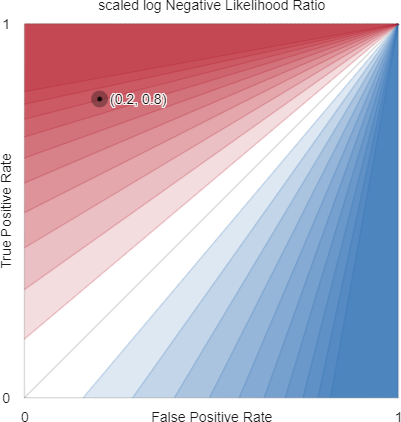}
  \label{fig:slogNLR}
\end{SCfigure}

\subsection{Negative Likelihood Ratio \texorpdfstring{($\LRN$)}{} and scaled log Negative Likelihood Ratio}
\label{sec:slLRN}


Using the notation of Table~\ref{tab:confusion-matrix}, the logarithm of the Likelihood Ratio of a negative outcomes  (Eq.~\eqref{eq:LRN}) can be rewritten in terms of $a, d, p, n$ as
\begin{align*}
    \log\LRN(a, b, c, d)
    &=\log\left(\frac{c}{p}\frac{n}{d}\right)\\
    &=\log\left(\frac{p-a}{p}\frac{n}{d}\right).
\end{align*}
For given numbers of positives ($p$) and negatives ($n$), this performance metric achieves a value of $ k $ along the contour lines with
\begin{align}
a(k, p, n, d) =  p - \frac{d e^k p}{n}
\label{eq:logLRN-contour}
\end{align}
or, in terms of true positive rate $\alpha=a/p$ and true negative rate $\delta=d/n$
\begin{align}
\alpha(k, \delta) = 
1 - \delta e^k
\end{align}

$\log\LRN$ has range $(-\infty,\infty)$. To visualise the finite values of this function, it is convenient to work with a scaled version of this function whose contours lie between $[-1,1]$.
The largest finite value of $\LRN$ is $pn/(n-1)$ so we can produce a scaled version of Eq.~\ref{eq:logLRN-contour} by dividing $\log\LRN(a, b, c, d)$ by
\begin{align}
M = \log(pn/(n-1))
\end{align}
to give
\begin{align}
     \text{scaled}\log\LRN(a, b, c, d)&=\frac{1}{\log(pn/(n-1))}\log\left(\frac{c}{p}\frac{n}{d}\right)
\end{align}
which yields the contour equations
\begin{align}
a(k, p, n, d) = p - \frac{d e^{Mk} p}{n}
\end{align}
and
\begin{align}
\alpha(k, \delta) = 
1 - \delta e^{Mk} 
\end{align}

\clearpage 

\begin{SCfigure}[50]
  \caption{Contours of Bookmaker Informedness in the ROC space.}
  \includegraphics[width=0.3\textwidth]{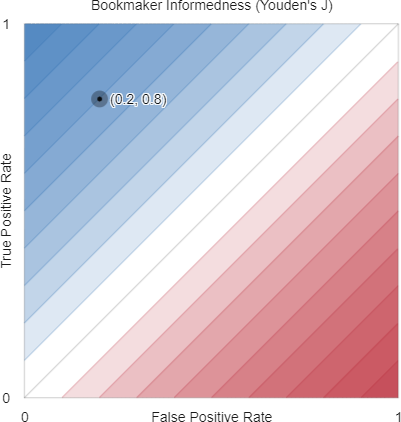}
  \label{fig:BM}
\end{SCfigure}

\subsection{Bookmaker Informedness, BA}
\label{sec:BA}


Using the notation of Table~\ref{tab:confusion-matrix}, the Bookmaker Informedness (Eq.~\eqref{eq:BM}) can be rewritten in terms of $a, d, p, n$ as
\begin{align*}
    \BM(a, b, c, d)
    &=\frac{a}{a+c} + \frac{d}{b+d} - 1\\
    &=\frac{a}{p} + \frac{d}{n} - 1.
\end{align*}
For given numbers of positives ($p$) and negatives ($n$), this performance metric achieves a value of $0 \leq k \leq 1$ along the contour lines with
\begin{align}
a(k, p, n, d) = p\left(k+1-\frac{d}{n}\right)
\end{align}
or, in terms of true positive rate $\alpha=a/p$ and true negative rate $\delta=d/n$
\begin{align}
\alpha(k, \delta) = 
k+1-\delta
\end{align}

\clearpage 

\begin{SCfigure}[50]
  \caption{Contours of Prevalence Threshold in the ROC space.}
  \includegraphics[width=0.3\textwidth]{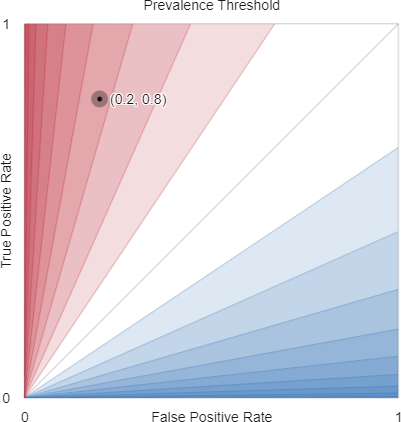}
  \label{fig:PT}
\end{SCfigure}

\subsection{Prevalence Threshold}


Using the notation of Table~\ref{tab:confusion-matrix}, the Prevalence Threshold (Eq.~\eqref{eq:PT}) can be rewritten in terms of $a, d, p, n$ as
\begin{align*}
    \PT(a, b, c, d)
    &=\frac{\sqrt{b/(b+d)}}{\sqrt{a/(a+c)} + \sqrt{b/(b+d)}}\\
    &=\frac{\sqrt{(n-d)/n}}{\sqrt{a/p} + \sqrt{(n-d)/n}}.
\end{align*}
For given numbers of positives ($p$) and negatives ($n$), this performance metric achieves a value of $0 \leq k \leq 1$ along the contour lines with
\begin{align}
a(k, p, n, d) = \frac{(k-1)^2 p (n-d}{k^2n}
\end{align}
or, in terms of true positive rate $\alpha=a/p$ and true negative rate $\delta=d/n$
\begin{align}
\alpha(k, \delta) = 
\frac{(k-1)^2  (1-\delta)}{k^2 }
\end{align}

\clearpage 

\begin{SCfigure}[50]
  \caption{Contours of scaled log Diagnostic Odds Ratio in the ROC space.}
  \includegraphics[width=0.3\textwidth]{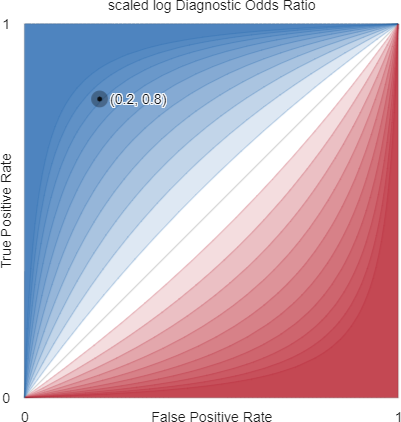}
  \label{fig:slogDOR}
\end{SCfigure}

\subsection{log Diagnostic Odds Ratio and scaled log Diagnostic Odds Ratio}
\label{sec:slDOR}

Using the notation of Table~\ref{tab:confusion-matrix}, the logarithm of the Diagnostic Odds Ratio (Eq.~\eqref{eq:DOR}) can be rewritten in terms of $a, d, p, n$ as
\begin{align}
    \log\DOR(a, b, c, d)
    &=\log\left(\frac{ad}{bc}\right)\\
    &=\log\left(\frac{ad}{(p-a)(n-d)}\right). \nonumber
\end{align}
For given numbers of positives ($p$) and negatives ($n$), this performance metric achieves a value of $k$ along the contour lines with
\begin{align}
a(k, p, n, d) = \frac{e^k(d - n) p}{d (e^k - 1 ) - e^k n}
\end{align}
or, in terms of true positive rate $\alpha=a/p$ and true negative rate $\delta=d/n$
\begin{align}
\alpha(k, \delta) = 
\frac{e^k  (\delta - 1) }{\delta (e^k - 1 ) - e^k } \label{eq:logDOR-contour}
\end{align}

$\log\DOR$ has range $(-\infty,\infty)$. To visualise the finite values of this function, it is convenient to work with a scaled version of this function whose contours lie between $[-1,1]$.
The largest finite value of $\DOR$ is $(p-1)(n-1)$ so we can produce a scaled version of Eq.~\ref{eq:logDOR-contour} by dividing $\log\DOR(a, b, c, d)$ by
\begin{align}
M = \log(p-1)(n-1)
\end{align}
to give
\begin{align}
     \text{scaled}\log\DOR(a, b, c, d)&=\frac{1}{\log(p-1)(n-1)}\log\left(\frac{ad}{bc}\right)
\end{align}
which yields the contour equations
\begin{align}
a(k, p, n, d) = \frac{e^{Mk}(d - n) p}{d (e^{Mk} - 1 ) - e^{Mk} n}
\end{align}
and
\begin{align}
\alpha(k, \delta) = 
\frac{e^{Mk}  (\delta - 1) }{\delta (e^{Mk} - 1 ) - e^{Mk} }.
\end{align}

\clearpage 

\begin{SCfigure}[50]
  \caption{Contours of the balanced Matthews Correlation Coefficient in the ROC space.}
  \includegraphics[width=0.3\textwidth]{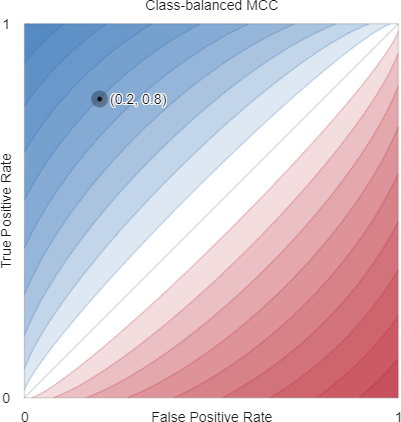}
  \label{fig:MCCbal}
\end{SCfigure}

\subsection{Balanced Matthews Correlation Coefficient}


Using the notation of Table~\ref{tab:confusion-matrix}, the balanced Matthews Correlation Coefficient (Eq.~\eqref{eq:bMCC}) can be rewritten in terms of $a, d, p, n$ as

\begin{align}
    \bMCC &= 
\frac
{a/p + d/n - 1}
{\sqrt{( 1 - \left(\frac{a}{p} \cdot \frac{d}{n} \right)^2 }}
\end{align}

For given numbers of positives ($p$) and negatives ($n$), this performance metric achieves a value of $k$ along the contour lines with
\begin{align}
a(k, p, n, d) = \frac{p}{n} \cdot \frac{k\sqrt{-4d^2 + 4dn + k^2 n^2} + n + d(k^2 - 1)}{k^2 + 1}
\end{align}
or, in terms of true positive rate $\alpha=a/p$ and true negative rate $\delta=d/n$
\begin{align}
\alpha(k, \delta) = 
\frac{k\sqrt{-4\delta^2 + 4d\delta + k^2 } + 1 + \delta(k^2 - 1)}{k^2 + 1}
\end{align}

\clearpage 

\begin{SCfigure}[50]
  \caption{Contours of the balanced Markedness in the ROC space.}
  \includegraphics[width=0.3\textwidth]{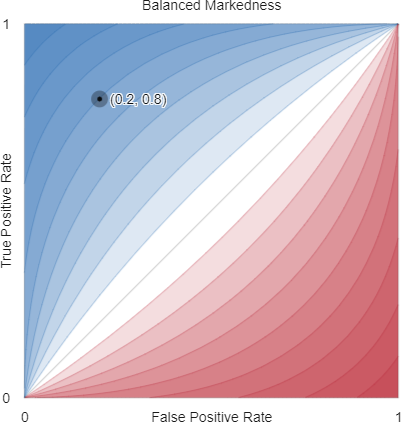}
  \label{fig:MKbal}
\end{SCfigure}

\subsection{Balanced Markedness}


Using the notation of Table~\ref{tab:confusion-matrix}, balanced Markedness (Eq.~\eqref{eq:bMK}) can be rewritten in terms of $a, d, p, n$ as

\begin{align}
    \bMK &= \frac{a/p}{a/p + 1 - d/n} + \frac{d/n}{d/n + 1 - a/p} - 1\\
         &= \frac{a}{a + p - pd/n} + \frac{d}{d + n - na/p} - 1
\end{align}

For given numbers of positives ($p$) and negatives ($n$), this performance metric achieves a value of $k$ along the contour lines with
\begin{align}
a(k, p, n, d) = 
p\left(\frac{d}{n} + \frac{\sqrt{(2k+1)^2 - 8dk/n} - 1}{2k}\right)
\end{align}
or, in terms of true positive rate $\alpha=a/p$ and true negative rate $\delta=d/n$
\begin{align}
\alpha(k, \delta) = 
\delta + \frac{\sqrt{(2k+1)^2 - 8\delta k} - 1}{2k}
\end{align}

\clearpage 

\begin{SCfigure}[50]
  \caption{Contours of $\bfOne$ in the ROC space.}
  \includegraphics[width=0.3\textwidth]{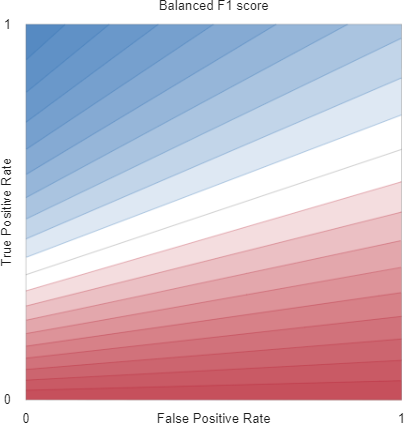}
  \label{fig:F1bal}
\end{SCfigure}

\subsection{Balanced \texorpdfstring{$\fOne$}{F1} \texorpdfstring{($\bfOne$)}{}}


Using the notation of Table~\ref{tab:confusion-matrix}, balanced $\fOne$ (Eq.~\eqref{eq:bF1}) can be rewritten in terms of $a, d, p, n$ as

\begin{align}
    \bfOne &= \frac{2a/p}{2 + a/p - d/n} 
\end{align}

For given numbers of positives ($p$) and negatives ($n$), this performance metric achieves a value of $k$ along the contour lines with
\begin{align}
a(k, p, n, d) = 
\frac{k p (d - 2 n)}{(k - 2) n} 
\end{align}
or, in terms of true positive rate $\alpha=a/p$ and true negative rate $\delta=d/n$
\begin{align}
\alpha(k, \delta) = 
\frac{k  (\delta - 2 )}{k - 2} 
\end{align}
in which case, all contours intersect at $(\alpha,\delta)=(0,2)$.

\clearpage 

\begin{SCfigure}[50]
  \caption{Contours of $\bFM$ in the ROC space.}
  \includegraphics[width=0.3\textwidth]{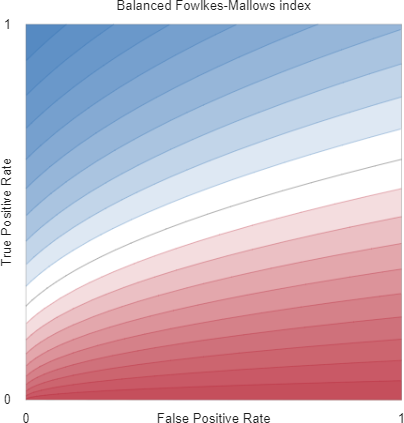}
  \label{fig:FMbal}
\end{SCfigure}

\subsection{Balanced Fowlkes-Mallows index}
\label{sec:balanced-FM}


Using the notation of Table~\ref{tab:confusion-matrix} and the same approach as \cite{luque_impact_2019}, a balanced version of the Fowlkes-Mallows Index  (Eq.~\eqref{eq:bFM}) can be rewritten in terms of $a, d, p, n$ as
\begin{align*}
    \bFM(a, b, c, d)
    &=\frac{\frac{a}{p}}{\sqrt{1+\frac{a}{p}-\frac{d}{n}}}
\end{align*}
For given numbers of positives ($p$) and negatives ($n$), this performance metric achieves a value of $0 \leq k \leq 1$ along the contour lines with
\begin{align}
a(k, p, n, d) = 
\frac{kp}{2} \left(\sqrt{k^2 + 4 - 4 d/n} + k\right)
\end{align}
or, in terms of true positive rate $\alpha=a/p$ and true negative rate $\delta=d/n$
\begin{align}
\alpha(k, \delta) = 
\frac{k}{2} \left(\sqrt{k^2 + 4 - 4 \delta} + k\right).
\end{align}

\clearpage 

\begin{SCfigure}[50]
  \caption{Contours of $\bTS$ in the ROC space.}
  \includegraphics[width=0.3\textwidth]{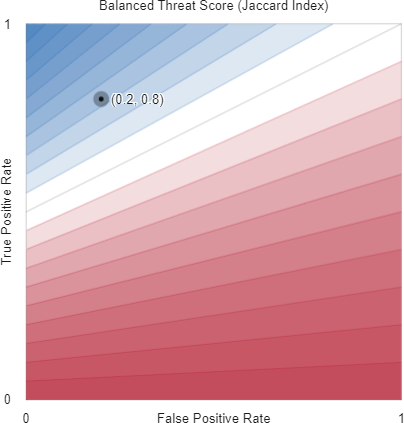}
  \label{fig:TSbal}
\end{SCfigure}

\subsection{Balanced Threat Score}
\label{sec:balanced-TS}


Using the notation of Table~\ref{tab:confusion-matrix} and the same approach as \cite{luque_impact_2019}, a balanced version of the Threat Score  (Eq.~\eqref{eq:bTS}) can be rewritten in terms of $a, d, p, n$ as
\begin{align*}
    \bTS(a, b, c, d)
    &=\frac{\frac{a}{p}}{2-\frac{d}{n}}
\end{align*}
For given numbers of positives ($p$) and negatives ($n$), this performance metric achieves a value of $0 \leq k \leq 1$ along the contour lines with
\begin{align}
a(k, p, n, d) = 
kp \left(2-\frac{d}{n} \right)
\end{align}
or, in terms of true positive rate $\alpha=a/p$ and true negative rate $\delta=d/n$
\begin{align}
\alpha(k, \delta) = 
k \left(2-\delta \right)
\end{align}

\clearpage 

\begin{SCfigure}[50]
  \caption{Contours of $\bPPV$ in the ROC space.}
  \includegraphics[width=0.3\textwidth]{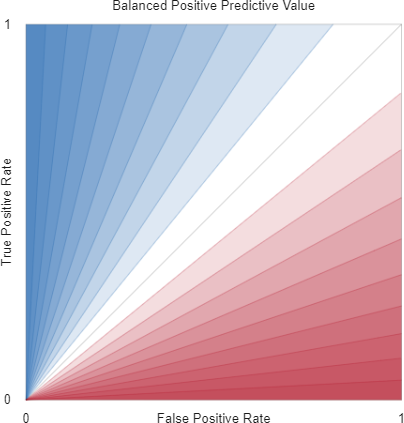}
  \label{fig:PPVbal}
\end{SCfigure}

\subsection{Balanced Positive Predictive Value}


Using the notation of Table~\ref{tab:confusion-matrix}, the balanced Positive Predictive Value  (Eq.~\eqref{eq:bPPV}) can be rewritten in terms of $a, d, p, n$ as
\begin{align*}
    \bPPV(a, b, c, d)
    &=\frac{\frac{a}{p}}{\frac{a}{p} + \left(1-\frac{d}{n}\right)}
\end{align*}
For given numbers of positives ($p$) and negatives ($n$), this performance metric achieves a value of $0 \leq k \leq 1$ along the contour lines with
\begin{align}
a(k, p, n, d) = 
\frac{k p (d - n)}{(k - 1) n}
\end{align}
or, in terms of true positive rate $\alpha=a/p$ and true negative rate $\delta=d/n$
\begin{align}
\alpha(k, \delta) = 
\frac{k  (\delta - 1)}{k - 1}.
\end{align}

\clearpage 

\begin{SCfigure}[50]
  \caption{Contours of $\bNPV$ in the ROC space.}
  \includegraphics[width=0.3\textwidth]{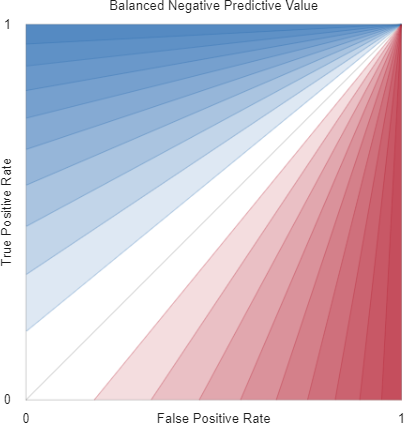}
  \label{fig:NPVbal}
\end{SCfigure}

\subsection{Balanced Negative Predictive Value}


Using the notation of Table~\ref{tab:confusion-matrix}, the balanced Negative Predictive Value  (Eq.~\eqref{eq:bNPV}) can be rewritten in terms of $a, d, p, n$ as
\begin{align*}
    \bNPV(a, b, c, d)
    &=\frac{\frac{d}{n}}{\frac{d}{n} + \left(1-\frac{a}{p}\right)}
\end{align*}
For given numbers of positives ($p$) and negatives ($n$), this performance metric achieves a value of $0 \leq k \leq 1$ along the contour lines with
\begin{align}
a(k, p, n, d) = 
\frac{dp(k-1)}{kn} + p
\end{align}
or, in terms of true positive rate $\alpha=a/p$ and true negative rate $\delta=d/n$
\begin{align}
\alpha(k, \delta) = 
\frac{\delta(k - 1)}{k} + 1.
\end{align}

\clearpage 

\subsection{Balanced Cohen's Kappa is Bookmaker Informedness}
\label{sec:CKbal}

Using the relationships:
\begin{align*}
    \TP &= p\cdot\TPR& \FP &= n(1-\TNR)\\
    \FN &= p(1-\TPR) & \TN &= n\cdot\TNR
\end{align*}
we can rewrite Cohen's Kappa (Eq.~\eqref{eq:CK}) as
\begin{align*}
\CK &= \frac
{2(p\cdot\TPR\cdot n\cdot\TNR - p(1-\TPR)n(1-\TNR))}
{(p\cdot\TPR+n(1-\TNR))n + p(p(1-\TPR)+n\cdot\TNR)}.
\end{align*}
Using the same approach as \cite{luque_impact_2019}, we can create a class-balanced version of this metric by setting $p=n$ to give an expression   in terms of the true positive and true negative rate alone:
\begin{align*}
\bCK &= \frac
{2(\TPR\cdot\TNR - (1-\TPR)(1-\TNR))}
{\TPR+1-\TNR + 1-\TPR+\TNR}\\
&= \frac
{2(\TPR\cdot\TNR - 1+\TPR+\TNR-\TPR\cdot\TNR)}
{2}\\
&=\TPR + \TNR - 1
\end{align*}
which is the same as Bookmaker Informedness (Eq.~\eqref{eq:BM}).

\clearpage

\section{Links to interactive visualisations, animations and figure source code}
\label{app:visualisations}

We have used R and Desmos' Graphing Calculator \citep{desmos_inc_desmos_nodate} to provide interactive visualisations for several key concepts in this paper. These visualisations are described in this section along with links to specific figures in the main paper. Desmos automatically ensures that the underlying code is available to copy and develop further and we provide RMarkdown for all other figures.

Source code (Rmarkdown) is available from Github at \url{https://github.com/DavidRLovell/Never-mind-the-metrics}.

\clearpage

\begin{SCfigure}[50]
  \caption{Screenshot of the interactive 3D confusion simplex (\url{http://bit.ly/see-confusion-simplex}).}
    \includegraphics[width=0.6\textwidth]{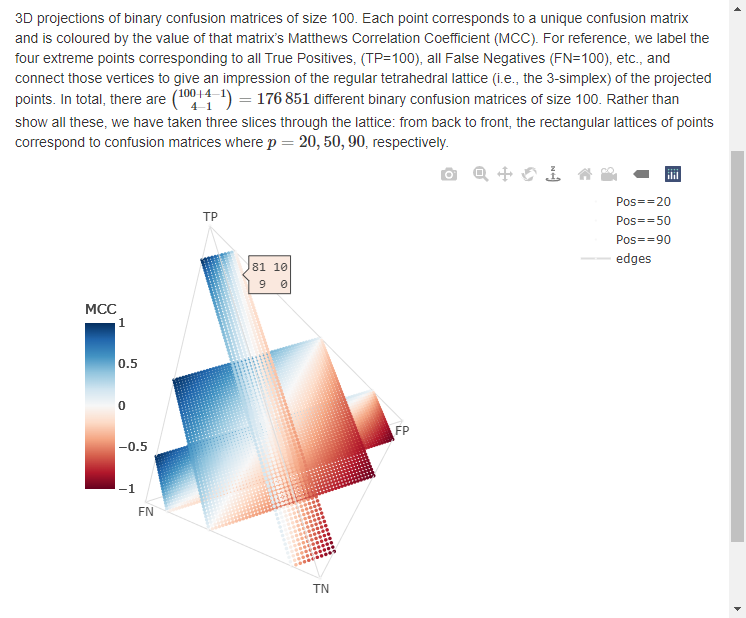}
  \label{fig:screenshot-confusion-simplex}
\end{SCfigure}

\subsection{Interactive 3D confusion simplex}
\label{app:simplex}

\url{http://bit.ly/see-confusion-simplex} shows an interactive visualisation of the 3D projection of binary confusion matrices of size 100. Each point corresponds to a unique confusion matrix and is coloured by the value of that matrix's Matthews Correlation Coefficient (MCC). Rather than show all $176\,851$ possible confusion matrices of size 100, we have taken three slices through the lattice: from back to front, the rectangular lattices of points correspond to confusion matrices where $p = 20, 50, 90$, respectively.

Users can mouse over the tetrahedron, then click and drag to change its orientation. Clicking on the text `Pos==20` will toggle that slice of the confusion matrix.

\clearpage

\begin{SCfigure}[50]
  \caption{Screenshot of the Desmos visualisation of possible points in ROC and Precision-Recall spaces (\url{http://bit.ly/see-ROC-reference-points}).}
    \includegraphics[width=0.6\textwidth]{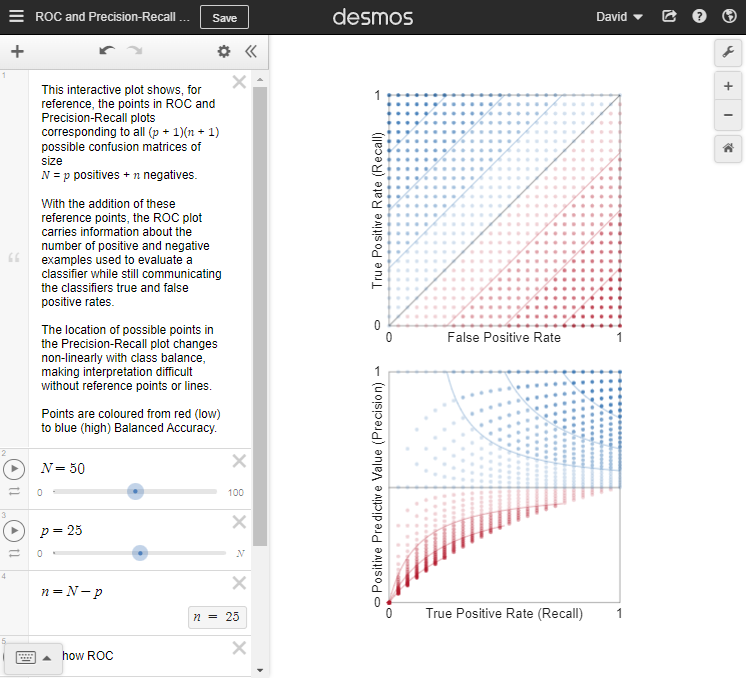}
  \label{fig:screenshot-ROC-PRC}
\end{SCfigure}

\subsection{All possible ROC and Precision-Recall reference points}
\label{app:ROC}

\url{http://bit.ly/see-ROC-reference-points} shows all possible $(p+1)\times(n+1)$ points in ROC and Precision-Recall spaces corresponding to confusion matrices of size $\Tot = \Pos + \Neg$, coloured from red (low) to blue (high) Balanced Accuracy. Users can change $\Tot$ and $\Pos$ by adjusting the sliders in the left hand side of the Desmos window. This visualisation was used to produce
\begin{itemize}
    \item Figure~\ref{fig:ROC.PRC.10}: see \url{http://bit.ly/ROC-PRC-10-40}
    \item Figure~\ref{fig:ROC.PRC.25}: see \url{http://bit.ly/ROC-PRC-25-25}
    \item Figure~\ref{fig:ROC.PRC.40}: see \url{http://bit.ly/ROC-PRC-40-10}
\end{itemize}

\clearpage

\begin{SCfigure}[50]
  \caption{Screenshot of the Desmos visualisation of confusion matrix performance metric contours (\url{http://bit.ly/see-confusion-metrics}). There are many things that users can switch on and off in this visualisation by clicking on the small round circles at the left edge of the screen.}
    \includegraphics[width=0.6\textwidth]{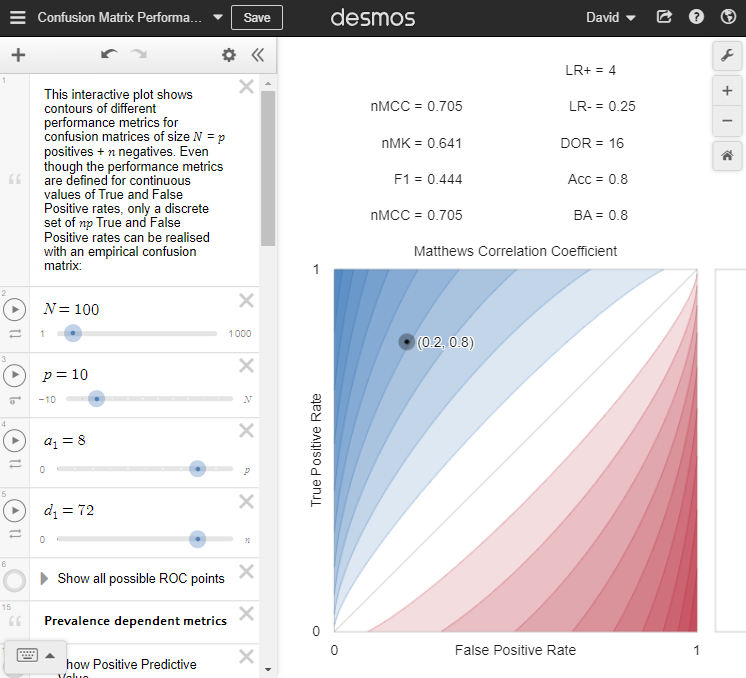}
  \label{fig:screenshot-contours}
\end{SCfigure}

\subsection{Confusion matrix performance metric contours}
\label{app:contours}

\url{http://bit.ly/see-confusion-metrics} enables us to interactively visualise a range of confusion matrix performance metrics by plotting their contours, coloured from red (low) to white (middle) to blue (high). This visualisation was used to produce all of the figures in Appendices~\ref{sec:prevalence-dependent-contours} and~\ref{sec:prevalence-independent-contours}.

Users can change $\Tot$ and $\Pos$ by adjusting the sliders in the left hand side of the Desmos window, and can set the position of a test point by adjusting the $a_1$ and $d_1$ sliders. There are many things that users can turn on and off by clicking on the small round circles at the left edge of the screen:
\begin{description}
      \item[Contours of prevalence-dependent and prevalence independent metrics.] These switches are titled \textsf{Show Accuracy}, \textsf{Show MCC}, through to  \textsf{Show Geometric Mean} and, when activated, display the contours of the chosen performance metrics
      \item[Additional information and decoration] switches allow users to show all possible ROC points; a movable test point whose corresponding confusion matrix and performance metric values can be displayed; and various titles. Importantly, users can toggle the limits of what is displayed, so that performance metric contours \textit{beyond} ROC space can be visualised (as in Figure~\ref{fig:MCCbeyond}).
\end{description}

\clearpage

\begin{SCfigure}[50]
  \caption{Screenshot of animation of Matthews Correlation Coefficient performance metric contours (\url{http://bit.ly/see-animated-MCC}).}
    \includegraphics[width=0.6\textwidth]{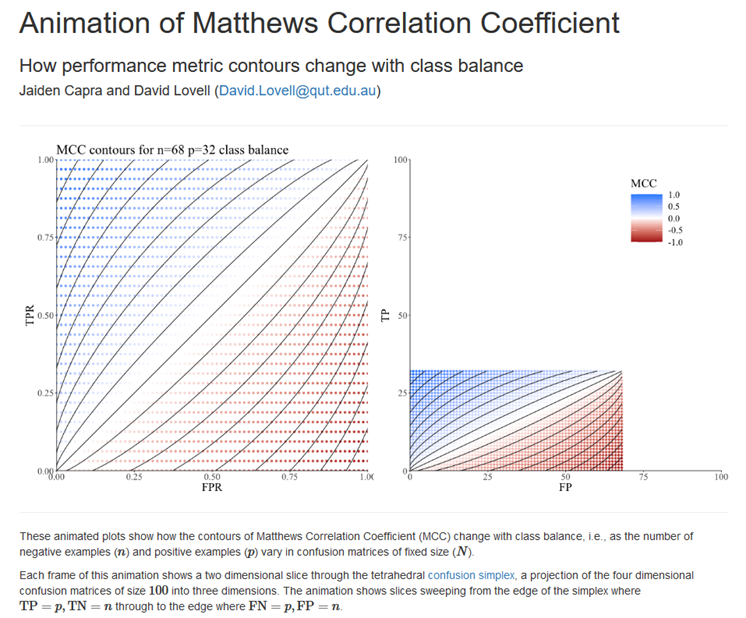}
  \label{fig:screenshot-MCC}
\end{SCfigure}

\subsection{Animated performance metric contours}
\label{app:animated-contours}

These animated plots show how the contours of various performance metrics change with class balance, i.e., as the number of negative examples ($n$) and positive examples ($p$) vary in confusion matrices of fixed size ($N$).

We have created animations of
\begin{itemize}
    \item Accuracy:                             \url{https://bit.ly/see-animated-accuracy}
    \item Balanced Accuracy:                    \url{https://bit.ly/see-animated-BA}
    \item $\fOne$ Score:                        \url{https://bit.ly/see-animated-F1}
    \item Matthews Correlation Coefficient:     \url{https://bit.ly/see-animated-MCC}
\end{itemize}

Each animation frame shows a two dimensional slice through the tetrahedral \href{http://bit.ly/see-confusion-simplex}{confusion simplex}, a projection of the four dimensional confusion matrices of size $100$ into three dimensions. The animation shows slices sweeping from the edge of the simplex where $\mathrm{TP}=p, \mathrm{TN}=n$ through to the edge where $\mathrm{FN}=p, \mathrm{FP}=n$.

Each coloured point corresponds to a specific confusion matrix in which 
$$
\begin{bmatrix}
\mathrm{TP} & \mathrm{FP}\\
\mathrm{FN} & \mathrm{TN}
\end{bmatrix}=
\begin{bmatrix}
\mathrm{TP} & \mathrm{FP}\\
p-\mathrm{TP} & n-\mathrm{FP}
\end{bmatrix}
$$
and $N=p+n=100$. Hence, for a given $p$ and $n$, we can plot the $(p+1)\times(n+1)$ points
whose $\mathrm{TP}$ values range from $0$ to $p$ and
whose $\mathrm{FP}$ values range from $0$ to $n$ while overlaying the contours of the `r metric.name` performance metric ranging from $-0.9, -0.8, \dots, 0.9$.

Note that

\begin{itemize}
    \item The contours of the performance metrics are defined continuously, but empirical confusion matrices can only take on values at the discrete points in these plots.
    \item The left hand plot shows these points and performance metric contours in ROC space in which  a classifier's true positive \textit{rate} is plotted against its false positive rate in the space of rational numbers from $[0,1]\times[0,1]$.
    \item The right hand plot shows these points and `r metric.name` contours as an orthographic projection of the slice of points from the confusion simplex.
    \item The left hand ROC plot is is equivalent to re-scaling the $x$-axis of the right hand plot by a factor of $\tfrac1n$ and the $y$-axis by $\tfrac1p$.
\end{itemize}

\clearpage

\begin{SCfigure}[50]
  \caption{Screenshot of the Desmos visualisation of confusion matrix uncertainty models (\url{http://bit.ly/see-confusion-uncertainty}). There are many things that users can switch on and off in this visualisation by clicking on the small round circles at the left edge of the screen.}
    \includegraphics[width=0.6\textwidth]{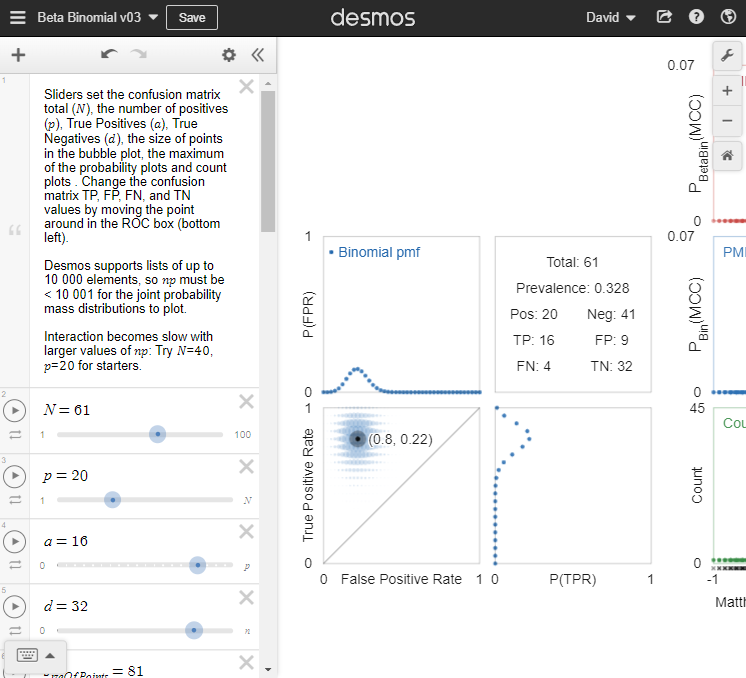}
  \label{fig:screenshot-Beta-Binomial}
\end{SCfigure}

\subsection{Uncertainty in confusion matrices and their performance metrics}
\label{app:uncertainty}

\url{http://bit.ly/see-confusion-uncertainty} enables interactive exploration of the posterior predictive pmfs of confusion matrices and three performance metrics (MCC, BA, $\fOne$) under binomial and beta-binomial models of uncertainty. This visualisation was used to produce Figures~\ref{fig:pmfs}, \ref{fig:pmfs-FN0}, \ref{fig:pmf.60} and~\ref{fig:pmf.61}.

Users can change $\Tot$ and $\Pos$ by adjusting the sliders in the left hand side of the Desmos window, and can set the position of a test point by adjusting the $a$ and $d$ sliders. There are many things that users can turn on and off by clicking on the small round circles at the left edge of the screen:
\begin{description}
      \item[Marginal and joint pmfs of True and False Positive rates.] Users can show these posterior predictive probability mass functions for confusion matrices of size $\Tot=\Pos+\Neg$ under binomial and beta-binomial models of uncertainty, given that $a$ True Positives and $d$ True Negatives have been observed.
      \item[Posterior predictive pmfs of MCC, BA and $\fOne$] can be shown using the \textsf{Show PMF...} switches for each performance metric. There are also switches to show the unique performance metric values (\textsf{Show rug...}), the number of times these unique values are observed (\textsf{Show count...}) and a histogram summary of the probability mass functions (\textsf{Show histogram...}).
      \item[Additional information and decoration] switches allow users to show all possible ROC points; a movable test point whose corresponding confusion matrix and performance metric values can be displayed; and various labels.
      \item[Axis and point size scales] are sliders that allow users to adjust the size of the points used in the joint pmf display, the maximum of the performance metric pmfs y-axis ($P_{max}$), and the maximum of the performance metric counts y-axis  ($C_{max}$).
\end{description}

As noted on the visualisation, Desmos supports lists of up to $10\,000$ elements, so $np$ must be $< 10\,001$ for the joint probability mass distributions to plot. This visualisation runs in your web browser and interaction becomes slow with larger values of $np$: we recommend starting with $N=60, p=20$.

\clearpage

\section{Who said ``the Matthews Correlation Coefficient is generally regarded as being a balanced measure''?}
\label{sec:MCCquote}

This quote is mentioned by \cite{zhu_performance_2020-1} but does not seem to come from either of the papers cited \citep{powers_evaluation_2011,chicco_ten_2017}. Google search for the phrase ``%
\href{https://www.google.com/search?q=%22Matthews+correlation+coefficient+is+generally+regarded+as%22}{Matthews Correlation Coefficient is generally regarded as}''
suggested it originated in Wikipedia and it can be found in the initial edit by user \texttt{JBom1} \cite{jbom1_matthews_2007} which says:
\begin{quote}
The Matthews Correlation Coefficient is used in machine learning as a measure of the quality of binary (two class) classifications. It takes into account true and false positives and negatives and is \textit{generally regarded as a balanced measure which can be used even if the classes are of very different sizes}. It returns a value between -1 and +1. A coefficient of +1 represents a perfect prediction, 0 an average random prediction and -1 the worst possible prediction.

While there is no perfect way of describing the confusion matrix of true and false positives and negatives by a single number, \textit{the Matthews Correlation Coefficient is generally regarded as being one of the best such measures}.

[Our emphasis]
\end{quote}

A later revision from IP Address \texttt{159.147.19.232} (\href{https://en.wikipedia.org/w/index.php?title=Matthews_correlation_coefficient&diff=prev&oldid=817508708}{Revision 28 December 2017}
attributes the statement 
to \citeauthor{boughorbel_optimal_2017}, though their paper says only ``We showed the suitability of MCC for imbalanced data'' \cite[Conclusions]{boughorbel_optimal_2017}.
A further revision by user \texttt{S-degoul} (\href{https://en.wikipedia.org/w/index.php?title=Matthews_correlation_coefficient&diff=next&oldid=666860409}{Revision 27 July 2015 }) attributes the second italicised statement in italics in the quote above to \cite{powers_evaluation_2011}, who says no such thing.

It seems to us that the assertions that Matthews Correlation Coefficient is generally regarded as a balanced measure, and as being one of the best performance measures currently reflects the opinions of three anonymous Wikipedia editors, rather than being statements backed up by evidence. If, however, people continue to quote these statements uncritically, we suspect they will become true in time.

\end{document}

%% file: newcommands.tex
\newcommand{\Tot}{N}
\newcommand{\Pos}{p}
\newcommand{\Neg}{n}

\newcommand{\PP}{\hat{p}}
\newcommand{\PN}{\hat{n}}
\newcommand{\LRP}{\mathrm{LR}_{+}}
\newcommand{\slLRP}{\mathrm{slLR}_{+}}
\newcommand{\LRN}{\mathrm{LR}_{-}}
\newcommand{\slLRN}{\mathrm{slLR}_{-}}
\newcommand{\TP}{\mathrm{TP}}
\newcommand{\FP}{\mathrm{FP}}
\newcommand{\TN}{\mathrm{TN}}
\newcommand{\FN}{\mathrm{FN}}
\newcommand{\TPR}{\mathrm{TPR}}
\newcommand{\FPR}{\mathrm{FPR}}
\newcommand{\TNR}{\mathrm{TNR}}
\newcommand{\FNR}{\mathrm{FNR}}
\newcommand{\DOR}{\mathrm{DOR}}
\newcommand{\slDOR}{\mathrm{slDOR}}
\newcommand{\BA}{\mathrm{BA}}
\newcommand{\BM}{\mathrm{BM}}
\newcommand{\Acc}{\mathrm{Acc}}
\newcommand{\DB}{\mathrm{DB}}
\newcommand{\MCC}{\mathrm{MCC}}
\newcommand{\bMCC}{\mathrm{MCC}_\mathit{bal}}
\newcommand{\Prev}{\mathrm{Prev}}
\newcommand{\PT}{\mathrm{PT}}
\newcommand{\MK}{\mathrm{MK}}
\newcommand{\bMK}{\mathrm{MK}_\mathit{bal}}
\newcommand{\PPV}{\mathrm{PPV}}
\newcommand{\bPPV}{\mathrm{PPV}_\mathit{bal}}
\newcommand{\NPV}{\mathrm{NPV}}
\newcommand{\bNPV}{\mathrm{NPV}_\mathit{bal}}
\newcommand{\fOne}{\mathrm{F}_1}
\newcommand{\bfOne}{\mathrm{F}_{1\mathit{bal}}}
\newcommand{\FM}{\mathrm{FM}}
\newcommand{\bFM}{\mathrm{FM}_\mathit{bal}}
\newcommand{\TS}{\mathrm{TS}}
\newcommand{\bTS}{\mathrm{TS}_\mathit{bal}}
\newcommand{\GM}{\mathrm{GM}}
\newcommand{\CK}{\kappa}
\newcommand{\bCK}{\kappa_\mathit{bal}}
\newcommand{\benefits}{\boldsymbol{\beta}}

%% file: table.binary.tex
\renewcommand\theadgape{\Gape[4pt]}
\renewcommand\cellgape{\Gape[4pt]}

\begin{tabular}{r|cc}
\multicolumn{1}{r|}{\textit{predicted}} &
\multicolumn{2}{c}{\textit{actual class}}                  \\
\multicolumn{1}{r|}{\textit{class}} &
\multicolumn{1}{c}{positive} &
\multicolumn{1}{c}{negative} \\ \hline
\multicolumn{1}{r|}{positive}           &
\makecell{\large{TP}\\ True\\ Positives\\ ($a$)}      &
\multicolumn{1}{c|}{\makecell{\large{FP}\\ False\\ Positives\\ ($b$)}}      \\
\multicolumn{1}{r|}{negative}           &
\makecell{\large{FN}\\ False\\ Negatives\\ ($c$)}      &
\multicolumn{1}{c|}{\makecell{\large{TN}\\ True\\ Negatives\\ ($d$)}}      \\ \cline{2-3} 
                  
\end{tabular}

%% file: eqn.statistics.tex

\newcommand{\MCCtext}{\pbox{9em}{\RaggedLeft Matthews Correlation\\Coefficient}}
\newcommand{\TStext}{\pbox{9em}{Jaccard index,\\ Critical Success index}}

The following equations define performance metrics and other quantities used in this paper. We use an asterisk ($^*$) to indicate quantities that depend on \textit{prevalence} (Equation~\eqref{eq:Prev}).

Here are the definitions of the row, column and overall totals of the confusion matrix:
\begin{flalign}
\text{Total             }        &  & \Tot  &= \TP  + \FN + \FP  + \TN                                              && \text{    }                         \label{eq:Tot} \\
\text{Condition Positive}        &  & \Pos  &= \TP  + \FN                                                           && \text{    }                         \label{eq:Pos} \\
\text{Condition Negative}        &  & \Neg  &= \FP  + \TN                                                           && \text{    }                         \label{eq:Neg} \\
\text{Predicted Positive}        &^*& \PP   &= \TP  + \FP                                                           && \text{    }                         \label{eq:PP}  \\
\text{Predicted Negative}        &^*& \PN   &= \FN  + \TN.                                                          && \text{    }                         \label{eq:PN}  \\
\intertext{\textit{Prevalence} refers to the proportion of positive cases in a dataset:}
\text{Prevalence}                &^*& \Prev &= \frac{\Pos}{\Tot}.                                                   && \text{    }                         \label{eq:Prev}\\
\intertext{True and False Positive rates form the Positive Likelihood Ratio:}
\text{True Positive Rate}        &  & \TPR  &= \frac{\TP}{\Pos} =    1 - \FNR                                       && \text{Sensitivity, Recall}          \label{eq:TPR} \\
\text{False Positive Rate}       &  & \FPR  &= \frac{\FP}{\Neg} =    1 - \TNR                                       && \text{    }                         \label{eq:FPR} \\
\text{Positive Likelihood Ratio} &  & \LRP  &= \frac{\TPR}{\FPR}                                                    && \text{    }                         \label{eq:LRP} \\
\intertext{while True and False Negative rates form the Negative Likelihood Ratio:}
\text{True Negative Rate}        &  & \TNR  &= \frac{\TN}{\Neg}                                                     && \text{Specificity}                  \label{eq:TNR} \\
\text{False Negative Rate}       &  & \FNR  &= \frac{\FN}{\Pos}                                                     && \text{    }                         \label{eq:FNR} \\
\text{Negative Likelihood Ratio} &  & \LRN  &= \frac{\FNR}{\TNR}                                                    && \text{    }                         \label{eq:LRN} \\
\intertext{and these Likelihood Ratios form the Diagnostic Odds Ratio:}
\text{Diagnostic Odds Ratio}     &  & \DOR  &= \frac{\LRP}{\LRN} = \frac{\TP\cdot\TN}{\FP\cdot\FN}.                 && \text{    }                         \label{eq:DOR} \\ 
\intertext{To help visualise $\LRP$, $\LRN$ and $\DOR$ in comparison to other performance metrics, we introduced the following scaled versions of their logarithms}
\text{scaled log $\LRP$} &  & \slLRP  &= \frac{1}{\log(n(p-1)/p)}\log\left(\frac{\TP}{p}\frac{n}{\FP}\right)        && \text{Section~\ref{sec:slLRP}}      \label{eq:slLRP}\\
\text{scaled log $\LRN$} &  & \slLRN  &= \frac{1}{\log(pn/(n-1))}\log\left(\frac{\FN}{p}\frac{n}{TN}\right)         && \text{Section~\ref{sec:slLRN}}      \label{eq:slLRN}\\
\text{scaled log $\DOR$} &  & \slDOR  &= \frac{1}{\log(p-1)(n-1)}\log\left(\frac{\TP\cdot\TN}{\FP\cdot\FN}\right)   && \text{Section~\ref{sec:slDOR}}      \label{eq:slDOR}\\
\intertext{True Positive and True Negative rates are the basis of the following prevalence-independent performance metrics:}
\text{Balanced Accuracy}         &  & \BA   &= \frac{\TPR+\TNR}{2}=\frac{\BM+1}{2}                                  && \text{    }                         \label{eq:BA}  \\ 
\text{Bookmaker Informedness}    &  & \BM   &= \TPR + \TNR - 1                                                      && \pbox{9em}{Youden's $J$,\\ Delta P} \label{eq:BM}  \\
\text{Geometric Mean}            &  & \GM   &= \sqrt{\TPR\cdot\TNR}                                                 && \text{    }                         \label{eq:GM}  \\
\text{Prevalence Threshold}      &  & \PT   &= \frac{\sqrt{\FPR}}{\sqrt{\TPR} + \sqrt{\FPR}}                        && \text{\cite{balayla_prevalence_2020}}\label{eq:PT}  \\ 
\intertext{Next come ratios that relate to a classifier's predictions. These depend on prevalence, but \cite{luque_impact_2019} have proposed prevalence-independent (``balanced'') versions:}
\text{Positive Predictive Value} &^*& \PPV  &= \frac{\TP}{\PP}                                                      && \text{Precision}                    \label{eq:PPV} \\ 
\text{Balanced PPV}              &  & \bPPV &= \frac{\TPR}{1+\TPR-\TNR}                                             && \text{\cite{luque_impact_2019}}     \label{eq:bPPV}\\ 
\text{Negative Predictive Value} &^*& \NPV  &= \frac{\TN}{\PN}                                                      && \text{\cite{powers_evaluation_2011}}\label{eq:NPV} \\ 
\text{Balanced NPV}              &  & \bNPV &= \frac{\TNR}{1+\TNR-\TPR}.                                            && \text{\cite{luque_impact_2019}}     \label{eq:bNPV}\\ 
\intertext{and these predictive values form Markedness and its balanced version:}
\text{Markedness}                &^*& \MK   &= \PPV + \NPV - 1                                                      && \text{    }                         \label{eq:MK}  \\ 
\text{Balanced Markedness}       &  & \bMK  &= \frac{\TPR}{1+\TPR-\TNR}+\frac{\TNR}{1+\TNR-\TPR}-1.                 && \text{\cite{luque_impact_2019}}     \label{eq:bMK} \\ 
\intertext{Accuracy refers to the proportion of correctly classified examples and is prevalence-dependent, unlike $\BA$ and $\BM$:}
\text{Accuracy}                  &^*& \Acc  &= \frac{\TP + \TN}{\Tot}.                                              && \text{    }                         \label{eq:Acc} \\ 
\intertext{Like Accuracy, both $\fOne$ and Threat Score involve ratios of sums of confusion matrix elements:}
\fOne                            &^*& \fOne &= \frac{2\PPV\cdot\TPR}{\PPV+\TPR}= \frac{2\TP}{2\TP+\FP+\FN}          && \pbox{9em}{Sørensen–Dice\\ coefficient}\label{eq:F1}\\  
\text{Balanced}~\fOne            &  & \bfOne&= \frac{2\TPR}{2+\TPR-\TNR}                                            && \text{\cite{luque_impact_2019}}     \label{eq:bF1} \\
\text{Threat Score}              &^*& \TS   &= \frac{\TP}{\TP+\FN+\FP}                                              && \TStext                             \label{eq:TS}  \\
\text{Balanced Threat Score}     &  & \bTS  &= \frac{\TPR}{2 - \TNR}.                                               && \text{Section~\ref{sec:balanced-TS}}\label{eq:bTS} \\
\intertext{We define Decision Benefits as a weighted sum of confusion matrix elements:}
\text{Decision Benefits}         &  & \DB   &= \beta_a\cdot\TP+\beta_b\cdot\FP+\beta_c\cdot\FN+\beta_d\cdot\TN.     && \text{Section~\ref{sec:decision-costs}}\label{eq:DB}\\
\intertext{The remaining metrics combine the four elements of the confusion matrix in more complex ways:}
\MCCtext                         &^*& \MCC  &= \frac{\TP\cdot\TN-\FP\cdot\FN}{\sqrt{\PP\cdot\Pos\cdot\Neg\cdot\PN}} && \text{phi coefficient}              \label{eq:MCC} \\ 
                                 &  &       &= \text{sgn}(\BM)\sqrt{\BM\cdot\MK}                                    && \text{\cite[Eq.19]{powers_evaluation_2011}}        \\ 
\text{Balanced MCC}              &  & \bMCC &= \frac{\BM}{\sqrt{1 - (\TPR-\TNR)^2}}                                 && \text{\cite{luque_impact_2019}}     \label{eq:bMCC}\\
\text{Fowlkes-Mallows Index}     &^*& \FM   &= \sqrt{\PPV\cdot\TPR}                                                 && \text{\cite{fowlkes_method_1983-1}} \label{eq:FM}  \\
\text{Balanced~}\FM              &  & \bFM  &= \frac{\TPR}{\sqrt{1 + \TPR - \TNR}}                                  && \text{Section~\ref{sec:balanced-FM}}\label{eq:bFM} \\
\text{Cohen's kappa}             &^*& \CK   &= \frac{2(\TP\cdot\TN-\FN\cdot\FP)}{\PP\cdot\Neg+\Pos\cdot\PN}.        && \text{See Section~\ref{sec:CKbal}}  \label{eq:CK}
\end{flalign}